\documentclass{article}
\usepackage{ijcai19}

\usepackage{times}
\usepackage{soul}
\usepackage{url}
\usepackage[hidelinks]{hyperref}
\usepackage[utf8]{inputenc}
\usepackage[small]{caption}
\usepackage{graphicx}
\usepackage{amsmath}
\usepackage{booktabs}
\usepackage{mathtools}
\usepackage{epsfig} 
\usepackage{amssymb}
\usepackage{sansmath}
\usepackage{amsfonts}

\usepackage{tabularx}
\usepackage{rotating}
\usepackage{booktabs}
\usepackage{multirow}
\usepackage{makecell}

\newcommand\norm[1]{\left\lVert#1\right\rVert}

\usepackage{xcolor}

\newcommand{\R}{\mathbb{R}}

\newcommand{\vvvPBoW}{\mathbf{v^{PBoW}}}
\newcommand{\vPBoW}{v^{PBoW}}
\newcommand{\vvvSPBoW}{\mathbf{v^{sPBoW}}}
\newcommand{\vSPBoW}{v^{sPBoW}}

\title{Persistence Bag-of-Words for Topological Data Analysis}

\author{
Bartosz Zieli\'nski$^1$\and
Micha\l{} Lipi\'nski$^1$\and
Mateusz Juda$^1$\and\\
Matthias Zeppelzauer$^2$\And
Pawe\l{} D\l{}otko$^3$
\affiliations
$^1$The Institute of Computer Science and Computer Mathematics,\\
Faculty of Mathematics and Computer Science, Jagiellonian University\\
$^2$Media Computing Group, Institute of Creative Media Technologies,\\
St. P\"{o}lten University of Applied Sciences,\\
$^3$Department of Mathematics and Swansea Academy of Advanced Computing,\\
Swansea University
\emails
\{bartosz.zielinski, michal.lipinski, mateusz.juda\}@uj.edu.pl,
m.zeppelzauer@fhstp.ac.at,
p.t.dlotko@swansea.ac.uk
}

\begin{document}

\maketitle

\begin{abstract}
Persistent homology (PH) is a rigorous mathematical theory that provides a robust descriptor of data in the form of persistence diagrams (PDs). PDs exhibit, however, complex structure and are difficult to integrate in today's machine learning workflows.
This paper introduces persistence bag-of-words: a novel and stable vectorized representation of PDs that enables the seamless integration with machine learning.
Comprehensive experiments show that the new representation achieves state-of-the-art performance and beyond in much less time than alternative approaches.
\end{abstract}

% ======================================================================
% ======================================================================
\section{Introduction}
\label{intro}

Topological data analysis (TDA) provides a powerful framework for the structural analysis of high-dimensional data. A main tool of TDA is Persistent Homology (PH)~\cite{edelsbrunner2010computational}, which currently gains increasing importance in data science~\cite{FerriSurvey2017}. It has been applied to a number of disciplines including, biology~\cite{protein_compressability}, material science~\cite{screening_of_materials}, analysis of financial markets~\cite{gidea_katz}. Persistence homology is also used as a novel measure of GANs (Generative Adversarial Networks) performance~\cite{pmlr-v80-khrulkov18a}, and as a complexity measure for neural network architectures~\cite{rieck2018neural}. PH can be efficiently computed using various currently available tools~\cite{bauer2017phat,Dey:2019:SET:3310279.3284360,maria2014gudhi}. A basic introduction to PH is given in the supplementary material (SM in the following)\footnotemark.

\footnotetext{Supplementary material: \url{https://arxiv.org/abs/1812.09245}}

The common output representation of PH are persistence diagrams (PDs) which are multisets of points in $\mathbb{R}^2$. Due to their variable size, PDs are not easy to integrate within common data analysis, statistics and machine learning workflows. To alleviate this problem, a number of kernel functions and vectorization methods for PDs have been introduced. 
Kernel-based approaches have a strong theoretical background but in practice they often become inefficient when the number of training samples is large. As the entire kernel matrix must usually be computed explicitly (like in case of SVMs), this leads to roughly quadratic complexity in computation time and memory with respect to the size of the training set. Furthermore, such approaches are limited to kernelized methods, such as SVM and kernel PCA.
Vectorized representations in contrast are compatible with a much wider range of methods and do not suffer from complexity constraints of kernels. Since they require a spatial quantization of the PD they might suffer from a loss in precision compared to kernels, especially since PDs are sparsely and unevenly populated structures.

In this work, we present a novel spatially adaptive and thus more accurate representation of PDs, which aims at combining the large representational power of kernel-based approaches with the general applicability of vectorized representations. To this end, we extend the popular bag-of-words (BoW) encoding (originating from text and image retrieval) to TDA to cope with the inherent sparsity of PDs~\cite{mccallum1998comparison,sivic2003video}. The proposed adaptation of BoW gives a universally applicable fixed-sized feature vector of low-dimension. It is, under mild conditions, stable with respect to a standard metric in PDs. Experiments demonstrate that our new representation achieves state-of-the-art performance and even outperforms numerous competitive methods while requiring orders of magnitude less time and being more compact. Due to the run-time efficiency of our approach it may in future enable the application of TDA for larger-scale data than possible today.

The paper is structured as follows. Section~\ref{background_and_rw} reviews related approaches. In Section~\ref{approaches} we introduce persistence bag-of-words and prove its stability. Sections~\ref{experiments} and \ref{results} present experimental setup, results and discussion. Please consider the SM$^a$ for additional information.

% ======================================================================
% ======================================================================
\section{Background and Related Work}
\label{background_and_rw}

Different kernel based and vectorized representations have been introduced to make PDs compatible with statistical analysis and machine learning methods. 
The goal of kernel-based approaches on PDs is to define dissimilarity measures (also known as kernel functions) used to compare PDs and thereby make them compatible with kernel-based machine learning methods like Support Vector Machines (SVMs) and kernel Principal Component Analysis (kPCA).
Li et al.~\cite{li2014persistence} use the traditional bag-of-features (BoF) approach combining various distances between $0$-dimensional PDs to generate kernels. On a number of datasets (SHREC 2010, TOSCA, hand gestures, Outex) they show that topological information is complementary to the information of traditional BoF.
Reininghaus et al.~\cite{reininghaus2014stable} turns PDs into a continuous distribution by appropriate placing of Gaussian distributions and use the scalar product of the distributions to define a kernel function.
Carri{\`e}re et al.~\cite{Carrire2017SlicedWK} propose a kernel based on sliced Wasserstein approximation of the Wasserstein distance.
Le and Yamada~\cite{le2018persistence} proposed a Persistence Fisher (PF) kernel for PDs. It has a number of desirable theoretical properties such as stability, infinite divisibility, and linear time complexity in the number of points in the PDs.
Lacombe et al.~\cite{lacombe2018large} reformulated the computation of diagram metrics as an optimal transport  problem. This approach allows for efficient parallelization and scalable computations of a PD's barycenters.
Another representation of PDs are Persistence Landscapes, PL~\cite{bubenik2015statistical}. PL is a transformation of PD into a sequence of piece-wise linear functions. $L^p$ distance between those functions or their scalar products can be used to define kernels.

Vectorized representations of PDs can be used directly as an input to most machine learning methods. 
Adams et al.~\cite{adams2017persistence}, propose the persistence image (PI) building upon earlier work~\cite{Frosini1998,Ferri1997}. PI first fits a distribution to the PD and then samples it at regular intervals to obtain a vectorized representation. Anirudh et al.~\cite{anirudh2016riemannian} propose an approach based on Riemannian manifold (RM). It transforms PD into a Gaussian kernel being a Riemannian manifold with Fisher-Rao metric and subsequently into a vector space which is reduced by PCA to a fixed-size representation.

Recently, a third type of approach has been introduced. It aims at learning areas/points in the PD which are of particular importance for a given task in a supervised manner~\cite{hofer2017deep}. Despite promising properties this approach can be used only in cases where supervisory information (labels) are available, requires a large training set and long training time. Additionally, it requires specifically adapted architectures for different data sets and data types~\cite{hofer2017deep}.

% ======================================================================
% ======================================================================
\section{Persistence Bag of Words}
\label{approaches}

In this section, we adopt the Bag-of-Words (BoW) model~\cite{mccallum1998comparison,sivic2003video}, introduced in text and image retrieval, for the quantization of PDs. The idea behind BoW is to quantize variable length input data into a fixed-size representation by a so called \textit{codebook}. The codebook is generated from the input data in an unsupervised manner by clustering. The basic assumption behind BoW is that the clusters (i.e. codewords) capture the intrinsic structure of the data and thereby represent a suitable vocabulary for the quantization of the data. 
Given a codebook $C$, every input point $P$ (in a potentially high-dimensional space) is encoded by assigning points from $P$ to the nearest codeword from $C$. In traditional BoW this encoding leads to a codeword histogram where each codeword from $C$ is a bin and counts how many points from $P$ are closest to it. 

For BoW approaches, three important hyperparameters need to be set: (1) the clustering algorithm used to generate the codebook, (2) the size of the codebook, i.e., the number of clusters, and (3) the type of proximity encoding which is used to obtain the final descriptors, i.e. hard or soft assignment. We employ k-means and Gaussian Mixture Models (GMM) for clustering. The size of each codebook is optimized to maximize performance. In the following sections, we will describe two ways of generating codeword histograms based on the proximity of points from $P$ to codewords in $C$.

\begin{figure*}[ht]
\centering
\includegraphics[width=0.99\linewidth]{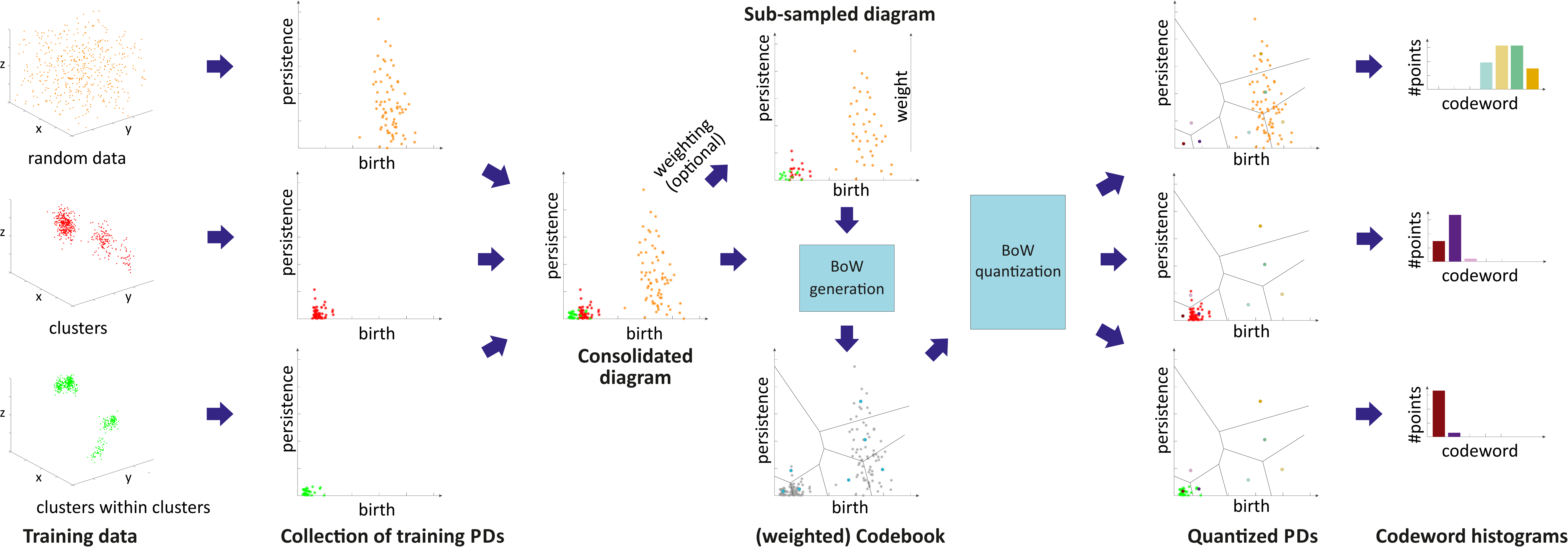}
\caption{Persistence bag-of-words: An illustration of the entire pipeline for codebook generation and the extraction of codeword histograms. From the input data we compute PDs in birth-persistence coordinates and combine them into one consolidated diagram. Next, a subset of points is obtained from this diagram by either weighted or unweighted sub-sampling. Subsequently, we cluster the sub-sampled consolidated diagram to derive a codebook. Finally, the individual points for each input PD are encoded by the codewords (BoW quantization). This illustration shows a hard assignment of points to codewords (for better illustration). In practice, a soft assignment strategy is recommended for stability reasons. The result is a codeword histogram for each input PD that represents how many points fall into which cluster of the codebook. These histograms represent a compact and fixed-size vectorial representation.}
\label{fig:approachPipeline}
\end{figure*}

The overall approach is visualized in Fig.~\ref{fig:approachPipeline}. The input consists of a set of PDs extracted from all instances of a given dataset and transformed into \emph{birth-persistence} coordinates using $(b,d) \rightarrow (b,d-b)$ transformation for each point in the PD. Next, all PDs are consolidated into one diagram which may then be sub-sampled to reduce the influence of noise. Based on the consolidated diagram, the codebook $C$ is generated by clustering. For a given codebook we generate codeword histograms by different quantization strategies. 

\subsection{Persistence Bag of Words}
\label{pbow}

In order to directly adapt BoW~\cite{baeza1999modern,sivic2003video} to PDs, let us consolidate a collection of training diagrams $D_1 , D_2, \ldots, D_n$ into $D = D_1 \cup D_2 \cup \ldots \cup D_n$ in order to obtain a codebook by using $k$-means clustering on $D$. Let $\{\mu_i \in \R^2, i=1,\ldots,N\}$ denote the centers of obtained clusters. Given a new PD $B=\{x_t \in \R^2\}_{t=1}^T$ let $NN(x_t)$ be equal to $i$ if $d(x_t,\mu_i) \leq d(x_t,\mu_j)$ for all $j \in \{1,\ldots,N\}$.
We define a \emph{persistence bag of words} (PBoW) as a vector:
\begin{equation}
\vvvPBoW(B) = \left(\vPBoW_i(B)\right)_{i=1,\ldots,N},
\end{equation}
where $\vPBoW_i(B)=card\{x_t\in B\ |\ NN(x_t)=i\}$. In other words, $\vPBoW_i$ captures the number of points from $B$, which are closer to $\mu_i$ than to any other $\mu_j$.  
Next, $\vvvPBoW(B)$ is normalized by taking the square root of each component (preserving the initial sign) and dividing it by the norm of the vector. This is a standard normalization for BoW~\cite{perronnin2010large} used to reduce the influence of outliers.

This direct adaptation of BoW to PDs is, however, not 1-Wasserstein stable. To show this, let us assume that we have two clusters with centers $\mu_1=(0,0),\mu_2 = (1,0)\in\R^2$, and PD $B$ containing only one point $x_1=(\frac{1}{2}+\epsilon,0)$, for some small $\epsilon>0$. 
Then, $\vvvPBoW(B)=[0, 1]$, because $x_1$ is closer to $\mu_2$ than $\mu_1$. However, a small perturbation in $B$, e.g. by $-2\epsilon$, changes the assignment of $x_1$ from $\mu_2$ to $\mu_1$. 
In this case $B'=\{ (\frac{1}{2}-\epsilon,0) \}$ and $\vvvPBoW(B')=[1, 0]$. 
In order to be stable in 1-Wasserstein sense, PBoW should fulfill the following condition:
\[
2=|\vvvPBoW(B)-\vvvPBoW(B'))|<C|x_1-y_1|<2C\epsilon,
\]
therefore $C>1/\epsilon$. As $\epsilon>0$ can be arbitrarily small, there does not exist a constant $C$ that meets this condition. Therefore PBoW is not stable. In the next section, we adopt BoW to better fit the structure of PDs and to deal with the instability.

% ==============================
\subsection{Stable Persistence Bag of Words}
\label{spbow}

In this section we present two important adaptations of BoW for PDs. Firstly, we enforce the codeword selection to be preferential to higher persistence points. Secondly, we adopt soft assignment of points to the clusters and prove that such an approach guarantees stability of the representation.

A consequence of the stability theorem for PDs is that points with higher persistence are typically considered more important than points with lower persistence. Therefore, when selecting the cluster centers $\mu_i$ in BoW, preference should be given to higher persistence points. To integrate this into codebook generation we perform the clustering on a subset of points obtained by a weighted sampling of $D$. For the sub-sampling we use a piece-wise linear weighting function $w_{a,b}: \R \rightarrow \R$. Given $a<b$:
\begin{equation}
w_{a,b}(t) = \left\{ \begin{array}{ll}
0 & \textrm{if $t < a$}\\
(t-a)/(b-a) & \textrm{if $a \leq t < b$}\\
1 & \textrm{if $b \leq t$}
\end{array} \right.
\end{equation}
and use it to weight the second coordinate (persistence) of the points in the PD. In our experiments we set $a$ and $b$ to the persistence values corresponding to $0.05$ and $0.95$ quantiles of the persistence coordinate of the points in $D$. Consequently, persistence points having higher values for function $w$ are more likely to be sampled. Note that the sub-sampling does not guarantee that the points of highest persistence will be selected as centers of clusters, but it makes the probability of such an event considerably larger. 

To account for the instability of PBoW (see Section~\ref{pbow}), we propose \emph{Stable Persistence Bag of Words} (sPBoW). Similarly to PBoW, we first consolidate PDs in the initial step of construction. Next, we generate a GMM based on the sub-sampled points (by expectation maximization~\cite{nasrabadi2007pattern}). This approach was originally introduced in \cite{van2008kernel}. Let the parameters of the fitted GMM be $\lambda=\{w_i,\mu_i,\Sigma_i, i=1,\ldots,N\}$, where $w_i$, $\mu_i$ and $\Sigma_i$ denote the weight, mean vector and covariance matrix of Gaussian $i$ and $N$ denotes the number of Gaussians. Given a PD $B$ its stable PBoW is defined as:
\begin{multline}
\vvvSPBoW(B)=\left(\vSPBoW_i=w_i\sum_{x_t\in B}p_i(x_t|\lambda)\right)_{i=1,\ldots,N},
\end{multline}
where $w_i>0$, $\sum_{i=1}^N w_i=1$, and $p_i(x_t|\lambda)$ is the likelihood that observation $x_t$ was generated by Gaussian $i$: 
\begin{multline}
p_i(x_t|\lambda) = \frac{exp\{-\frac{1}{2}(x_t-\mu_i)'\Sigma_i^{-1}(x_t-\mu_i)\}}{2\pi|\Sigma_i|^{\frac{1}{2}}}.\nonumber
\end{multline}

See SM$^a$ for a complexity analysis of PBoW and sPBoW.

% ======================================================================
% ======================================================================
\section{Stability Proof}
\label{sec:stable_spbow}

\medskip
\noindent
{\bf Theorem.} Let $B$ and $B'$ be persistence diagrams with a finite number of non-diagonal points. Stable persistence bag of words, sPBoW with $N$ words is stable with respect to \mbox{1-Wasserstein} distance between the diagrams, that is
\[
\norm{\vvvSPBoW(B)-\vvvSPBoW(B')}_{\infty}\leq C\cdot W_1(B, B'),
\]
where $C$ is a constant.

\medskip
\noindent
{\it Proof}. 
Let $\eta : B \rightarrow B'$ be an optimal matching in the definition of 1-Wasserstein distance. For a fixed $i\in\{1,\ldots,N\}$ we have:

\begin{multline}
\norm{\vSPBoW_i(B)-\vSPBoW_i(B')}_\infty= \\
\norm{w_i\sum_{x \in B}( p_i(x|\lambda)- p_i(\eta(x)|\lambda)}_\infty \leq \\
|w_i|\sum_{x \in B}\norm{\left(p_i(x|\lambda)-p_i(\eta(x)|\lambda)\right)}_\infty\nonumber
\end{multline}
Note that $p_i: \R^2 \rightarrow \R$ are Lipschitz continuous (as they are 2-dimensional Gaussian distributions). Let $L_i$ be their Lipschitz constant. We get
\begin{multline}
|w_i|\sum_{x \in B}\norm{\left(p_i(x|\lambda)-p_i(\eta(x)|\lambda)\right)}_\infty \leq \\
|w_i|\sum_{x \in B}\norm{\left( L_i(x - \eta(x))\right)}_\infty = \\ 
|w_i\ L_i|\sum_{x \in B}\norm{\left( x - \eta(x)\right)}_\infty = |w_i\ L_i|\ W_1(B, B')\nonumber
\end{multline}
\noindent Consequently for $C = max_{i} |w_i\ L_i|$ we have
\[
\norm{\vvvSPBoW(B)-\vvvSPBoW(B')}_{\infty} \leq C\ W_1(B, B').
\]
\begin{flushright} 
$\square$
\end{flushright}

% ======================================================================
% ======================================================================
\section{Experimental Setup}
\label{experiments}

% ==============================
\subsection{Datasets}
\label{subsec:datasets}

We incorporate datasets which cover a wide range of different retrieval problems. Firstly, to provide a proof-of-concept, we evaluate all approaches on a set of synthetically generated shape classes from~\cite{adams2017persistence}. It consists of six shape classes represented by point clouds of the geometrical objects. The task is to differentiate them by the derived representations. Additionally, we evaluate the approaches on real-world datasets for geometry-informed material recognition (GeoMat)~\cite{degol2016geometry}, classification of social network graphs (reddit-5k, reddit-12k)~\cite{hofer2017deep}, analysis of 3D surface texture (PetroSurf3D)~\cite{zeppelzauer2017study}, and 3D shape segmentation~\cite{Carrire2017SlicedWK}. Where available, we have used pre-computed PDs available with datasets to foster reproducibility and comparability. More detail on the datasets is provided in the SM$^a$.

% ==============================
\subsection{Compared Approaches}

We compare our bag-of-word approaches with both kernel-based techniques and vectorized representations. Kernel-based approaches include: 2-Wasserstein distance\footnote{\url{https://bitbucket.org/grey_narn/hera}} (2Wd)~\cite{kerber2017geometry}, the multi-scale kernel\footnote{\url{https://github.com/rkwitt/persistence-learning}} (MK) of~\cite{reininghaus2014stable}, and sliced Wasserstein kernel\footnote{code obtained from Mathieu Carri{\`e}re} (SWK)~\cite{Carrire2017SlicedWK}. Furthermore, we employ the persistence landscape\footnote{\url{https://www.math.upenn.edu/~dlotko}}$^,$\footnote{\url{https://github.com/queenBNE/Persistent-Landscape-Wrapper}} (PL) representation and generate a kernel matrix by the distance metric defined in~\cite{bubenik2015statistical}. Vectorized PD representations include: persistence image\footnote{\url{https://github.com/CSU-TDA/PersistenceImages}} (PI)~\cite{adams2017persistence} and the Riemannian manifold approach\footnote{\url{https://github.com/rushilanirudh/pdsphere}} (RM)~\cite{anirudh2016riemannian}. 

% ==============================
\subsection{Setup}
\label{subsec:setupOfExperiments}

For all datasets, we aim at solving a supervised classification task. The classification pipeline is as follows: for the kernel-based approaches we take the PDs as input and compute the kernel matrices for the training and test samples. Next we train an SVM from the explicit kernel matrices and evaluate it on the test data. For the vectorized representations we compute the respective feature vectors from the PDs and feed them into a linear SVM for training. This procedure allows for directly comparing kernel-based approaches and vectorized representations. To enhance the comparability we employ (if available) the original train/test division of the datasets. To find optimal parameters for each evaluated approach, we run a grid search including cross-validation over the hyperparameters of all approaches (see Table 1 in SM$^a$).

As the computation times for some of the considered methods, especially for kernel-based approaches, do not scale well with the sizes of datasets (computation time and space required for explicit kernel matrices grows exponentially), we have decided to split the evaluation into two parts: EXP-A uses all related approaches on smaller randomly sub-sampled versions of the datasets (see SM$^a$ for details on sub-sampling), while EXP-B operates only on the vectorized representations and uses larger datasets.

The mean accuracy is obtained as an average over 5 runs with the same train/test divisions used by the compared methods. A Wilcoxon signed-rank tests is used to show the statistical significance of differences between the scores of the best and the remaining methods. As the number of repetitions is small, the p-value is set to 0.1 and no multiple testing correction is applied.

The code of our experiments is implemented in Matlab\footnote{\url{https://github.com/bziiuj/pcodebooks}}. For external approaches we use the publicly available implementations of the original authors. For bag-of-words we employ the VLFeat library~\cite{vedaldi08vlfeat}.

% ======================================================================
% ======================================================================

\section{Results and Discussion}
\label{results}

Table~\ref{tab:kernelResults} summarizes the results obtained in our experiments for EXP-A and EXP-B. For each combination of dataset and approach we provide the obtained classification accuracy (including the standard deviation) and the processing time needed to construct the representations (excluding the time for classification). Note that for the synthetic dataset and the 3D shape segmentation dataset, results of EXP-A and EXP-B are equal, as no sub-sampling was needed to perform EXP-A.

\setlength{\tabcolsep}{4pt}
\begin{table*}[ht]
\begin{center}
\begingroup
\fontsize{5.5pt}{5.5pt}\selectfont
\begin{tabular}{c|rr|rr|rr|rr|rr|rr}
\toprule
\multicolumn{13}{c}{\thead{{\bf EXP-A}}} \\
{\bf Descr.} & 
\multicolumn{2}{c}{{\bf Synthetic}} &
\multicolumn{2}{|c}{{\bf GeoMat}} & 
\multicolumn{2}{|c}{{\bf Reddit-5k}} &
\multicolumn{2}{|c}{{\bf Reddit-12k}} &
\multicolumn{2}{|c}{{\bf PetroSurf3D}} &
\multicolumn{2}{|c}{{\bf 3D Shape Segm.}}\\
{} & Score & Time (sec.) & Score & Time (sec.) & Score & Time (sec.) & Score & Time (sec.) & Score & Time (sec.) & Score & Time (sec.) \\
\hline
2Wd & 
    \boldmath$98.0\pm1.3$ & $133.4$ & 
    \boldmath$24.7\pm3.1$ & $14740.6$ &
    $29.6\pm5.6$ & $6198.8$ &
    $23.3\pm2.4$ & $4979.8$ &
    \boldmath$79.9\pm6.6$ & $63822.5$ &
	$71.0\pm0.6$ & $6300.8$
    \\
MK & 
    $92.3\pm2.8$ & $44.0$ & 
    $9.1\pm3.1$ & $10883.0$ & 
    $32.7\pm3.1$ & $7251.7$ & 
    $26.7\pm4.6$ & $4222.3$ & 
    \boldmath$80.2\pm3.6$ & $53831.0$ & 
    $88.5\pm0.5$ & $860.3$ \\
SWK & 
    \boldmath$97.4\pm1.9$ & $33.3$ & 
    $22.0\pm2.7$ & $1069.6$ &
    $39.2\pm6.1$ & $373.8$ &
    $30.9\pm5.0$ & $329.7$ &
    \boldmath$78.0\pm6.3$ & $3913.0$ &
	$94.2\pm0.5$ & $1995.4$
    \\
PL & 
    $95.1\pm2.2$ & $81.0$ & 
    $17.9\pm3.8$ & $1563.7$ &
    $30.4\pm6.2$ & $791.3$ &
    $24.7\pm4.2$ & $719.0$ &
    \boldmath$78.2\pm6.7$ & $7991.7$ &
	$92.6\pm0.8$ & $2668.3$
    \\
\hline
PI & 
    \boldmath$98.3\pm1.6$ & $10.1$ & 
	$11.1\pm2.4$ & $342.3$ &
	\boldmath$48.4\pm5.7$ & $1250.9$ &
    \boldmath$35.6\pm4.6$ & $111.8$ &
    \boldmath$80.0\pm3.7$ & $1877.0$ &
	\boldmath$94.9\pm0.3$ & $291.2$
    \\
RM & 
    $92.3\pm2.6$ & \boldmath$0.9$ & 
	$19.1\pm2.7$ & $10.3$ &
	$39.2\pm9.2$ & $7.8$ &
    $28.7\pm2.0$ & $17.9$ &
    \boldmath$77.3\pm6.3$ & $128.8$ &
	$72.3\pm0.4$ & $6.6$
    \\
PBoW  & 
    \boldmath$98.0\pm2.2$ & \boldmath$0.8$ & 
	\boldmath$26.7\pm2.5$ & \boldmath$0.3$ &
	\boldmath$46.8\pm4.8$ & \boldmath$0.5$ &
    \boldmath$32.7\pm2.2$ & \boldmath$0.3$ &
    \boldmath$79.9\pm4.6$ & \boldmath$1.8$ &
	$90.8\pm0.7$ & \boldmath$5.4$
    \\
sPBoW &
    \boldmath$96.9\pm1.6$ & $4.0$ & 
	$22.2\pm2.6$ & $2.2$ &
	\boldmath$45.6\pm5.4$ & $0.7$ &
    $31.6\pm2.8$ & \boldmath$1.3$ &
    \boldmath$78.8\pm3.7$ & $31.5$ &
	\boldmath$94.4\pm0.6$ & $14.7$
    \\
\midrule
\multicolumn{13}{c}{\thead{{\bf EXP-B}}} \\
% {\bf Descr.} & 
% \multicolumn{2}{c}{{\bf Synthetic}} &
% \multicolumn{2}{|c}{{\bf GeoMat}} & 
% \multicolumn{2}{|c}{{\bf Reddit-5k}} &
% \multicolumn{2}{|c}{{\bf Reddit-12k}} &
% \multicolumn{2}{|c}{{\bf PetroSurf3D}} &
% \multicolumn{2}{|c}{{\bf 3D Shape Segm.}}\\
% {} & Score & Time (sec.) & Score & Time (sec.) & Score & Time (sec.) & Score & Time (sec.) & Score & Time (sec.) & Score & Time (sec.) \\
\hline
PI & 
    \multicolumn{2}{c|}{\multirow{4}{*}{same as EXP-A}} & 
    $22.45\pm0.0$ & $7243.1$ &
    \boldmath$49.3\pm2.8$ & $4758.7$ &
    \boldmath$38.4\pm0.9$ & $11329.0$ &
    \boldmath$80.4\pm4.9$ & $12137.0$ &
    \multicolumn{2}{c}{\multirow{4}{*}{same as EXP-A}} \\
RM & {} & {} & 
	$9.4\pm0.0$ & $222.6$ &
    $46.3\pm3.1$ & $108.1$ &
    $32.6\pm1.1$ & $215.5$ &
    \boldmath$79.9\pm5.0$ & $1451.0$ &
    {} & {} \\
PBoW & {} & {} & 
    \boldmath$29.0\pm0.4$ & \boldmath $5.2$ &
    \boldmath$49.9\pm3.3$ & \boldmath$1.5$ &
    \boldmath$38.6\pm0.9$ & \boldmath$4.8$ &
    \boldmath$80.3\pm5.3$ & \boldmath$28.5$ &
    {} & {} \\
sPBoW & {} & {} & 
    $27.8\pm0.7$ & $29.7$ &
	$46.6\pm3.3$ & $5.2$ &
    $35.3\pm1.4$ & $29.3$ &
    \boldmath$79.9\pm5.2$ & $161.8$ &
    {} & {} \\
% \bottomrule
% \hline
\hline
SoA & 
    \multicolumn{2}{c|}{$97.3$ \cite{adams2017persistence}} &
    \multicolumn{2}{c|}{$22.3\pm0.8$ \cite{degol2016geometry}} &
    \multicolumn{2}{c|}{$49.1$ \cite{hofer2017deep}} &
    \multicolumn{2}{c|}{$38.5$ \cite{hofer2017deep}} &
    \multicolumn{2}{c|}{n/a} &
    n/a & \\ 
\midrule
\end{tabular}
\caption{Results of EXP-A and EXP-B averaged over 5 runs. We provide average classification accuracy, standard deviation and computation time. Results with statistically significant improved performance are highlighted bold. The first four approaches are kernels (2Wd: 2-Wasserstein Distance, MK: Multiscale Kernel, SWK: Sliced Wasserstein Kernel, PL: Persistence Landscape). The remaining approaches are vectorized representations evaluated in EXP-A and EXP-B (PI: Persistence Image, RM - Riemmanian Manifold, (s)PBoW: (stable) Persistence Bag-of-Words). Column ``Time" shows the computation times for kernel matrices or vectorized representations using the optimal parameters obtained in grid search (excluding cross-validation and classification). Times highly depend on those parameters (e.g. the resolution of PI), which is further investigated in Section~\ref{subsec:resultAccuracyvsCodebookSize} and in the SM$^a$. Times (esp. in EXP-A) further vary significantly with the dataset that influences the number of points in the PDs. Row "SoA" shows comparable state-of-the-art results from the literature where available.}
\label{tab:kernelResults}
\endgroup
\end{center}
\end{table*}

Overall, for all experiments in EXP-A and EXP-B, PBoW or sPBoW achieve state-of-the-art performance or even above. From EXP-A we further observe that vectorized representations (including the proposed ones) in all cases outperform kernel-based approaches. Among the compared vectorized representations, PI in most cases outperforms RM and will thus serve as the primary approach for further comparison. When comparing the stable vs. unstable variants of PBoW, we observe that PBoW in most cases outperforms its stable equivalent. This is further studied in Section~\ref{subsec:resultAccuracyvsCodebookSize}.

Large differences exist in processing times of the different approaches. The slowest vectorized approach is PI. The runtimes, however, vary strongly, depending on the resolution used for PI, which depends on the optimally estimated resolution during grid search. The RM representation is one to two magnitudes faster than PI\footnote{For both representations the original implementations are used}. All kernel methods are about one magnitude or more slower then PI.
(s)PBoW outperform almost all state of the art approaches in runtime for all datasets (in EXP-A and EXP-B). The gain in runtime efficiency ranges from one to up to four orders of magnitude. For the largest dataset (PetrSurf3D), for example, the PBoW and sPBoW require $29$ and $162$ seconds while RM requires 1.451 seconds and PI 12.136 seconds. In the following, we analyze selected aspects of the proposed representations in greater detail.

% ==============================
\subsection{Accuracy vs. Codebook Size}
\label{subsec:resultAccuracyvsCodebookSize}

\begin{figure}[ht]
\centering
\includegraphics[width=\linewidth]{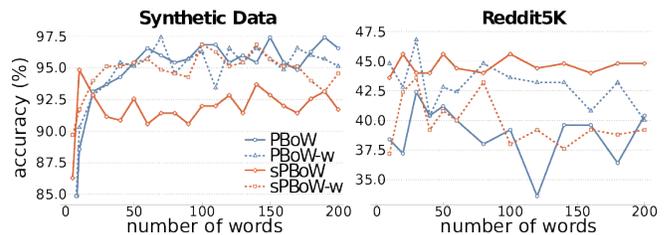}
\caption{Accuracy vs. codebook size for datasets from EXP-A without (solid lines) and with codebook weighting (dashed lines). See plots for the remaining datasets in Fig.~3 of the SM$^a$.}
\label{fig:acc_words_exp1}
\end{figure}

The most important parameter for BoW representations is the codebook size $N$, i.e. the number of clusters. There is no commonly agreed analytic method to estimate the optimal codebook size, thus the estimation is usually performed empirically. To investigate the sensitivity of (s)PBoW and their performance on the codebook size, we evaluate both approaches for a range of values of $N$ for each dataset. Results are shown in Fig.~\ref{fig:acc_words_exp1}, both without (solid lines) and with weighted sub-sampling (dashed lines) of the consolidated PD.

On the synthetic dataset we observe that without weighting PBoW outperforms sPBoW. However, when codebook weighting is introduced, the stable variant sPBoW starts to work equally well. The trend shows that larger codebook sizes are better than small ones but also that with already 20 words a high performance can already be achieved. Using weighting and the stable formulation of PBoW clearly improves the performance on this dataset. 

Unlike to the synthetic data, in the case of social network graphs (reddit-5k) there is no recognizable increase in performance for larger codebook sizes. We assume that this is caused by the larger variation between training and test data in real data sets. More precisely, larger codebooks result in codewords which tend to overfit on the training PDs. Such codewords do not generalize well to test data, which can result in the observed behavior.

Over all datasets we observe that the benefit of weighting is dataset dependent and that no general recommendation can be derived (see results for the remaining datasets in SM$^a$). The stable formulation of PBoW, however, performs equally well or even better than the unstable one in 4/6 datasets.

% ==============================
\subsection{Qualitative Analysis}
\label{subsec:resultsPBoWforSyntheticDataset}

\begin{figure}[ht]
\begin{center}
\includegraphics[width=\linewidth]{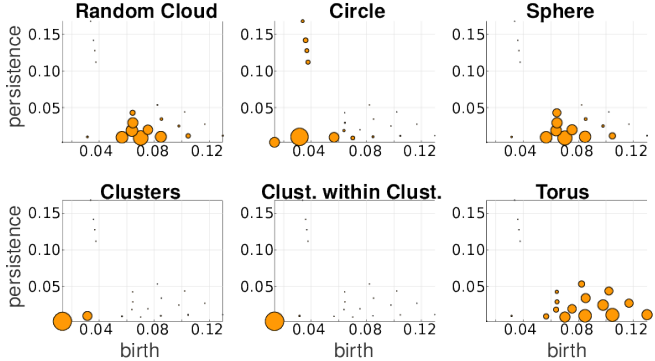}
\caption{Average codebook histograms computed for each of the six shape classes of the synthetic dataset. The cluster center of each codeword is presented as a circle in the birth-persistence domain. The area of the circles reflects the histogram values of the specific class. For all classes the same codebook (same clustering) is employed, thus, dot locations are the same on all plots. Class differences are thus reflected by different sizes of the circles.}
\label{fig:bow_assignment}
\end{center}
\end{figure}

In this section we investigate the proposed representations with a special focus on their discriminative abilities. We employ the synthetic dataset as proof-of-concept and GeoMat as example of a complex real-world dataset. We compute PBoW with $N=20$ clusters for the synthetic dataset and visually analyze the codeword histograms obtained by (hard) assignment. For each of the six shape classes we compute the average codebook histogram (over all samples of each class) to obtain one representative PBoW vector per class. The averaged PBoW histograms for each classes are presented in Fig.~\ref{fig:bow_assignment}. Instead of only providing the histograms themselves, we plot for each codeword of the histogram the corresponding cluster center as a circle in the original birth-persistence domain and encode the number of assigned codeworks (the actual values of the histograms) in the area of the circles, i.e. the larger the count for a cluster, the larger the circle. The advantage of this representation is that the spatial distribution of the codewords in the PD is preserved.

From Fig.~\ref{fig:bow_assignment} we can see that except for the classes ``random cloud" and ``sphere" (which are difficult to differentiate) all classes generate strongly different cluster distributions. Class ``circle", for example, uniquely activates four clusters with strong persistence (top-left corner) and the ``torus" class distributes its corresponding code words across a large number of clusters representing less persistent components. 

Fig.~\ref{fig:bow_assignment} further illustrates an important property of persistence bag-of-words, namely its sparse nature. More specifically, areas with no points in the consolidated persistence diagram will contain no codewords (clusters). In Fig.~\ref{fig:bow_assignment}, for example, no codeword is obtained in the upper-right quadrant of the diagram, since no components are located there for the underlying data. Thus these unimportant areas are neglected and not encoded into the final representation. This not only reduces the dimension of the final representation but further makes the representation adaptive to the underlying data and increases the information density of the representation.

\begin{figure}
\begin{center}
\includegraphics[width=\linewidth]{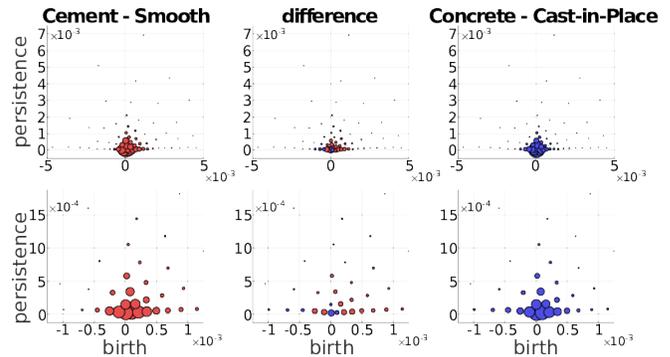}
\caption{Comparison of averaged PBoW histograms for class ``cement smooth'' (left, red) and ``concrete cast-in-place'' (right, blue) from GeoMat (top row: total view, bottom row is zoomed in). The plot in the center shows the difference between the classes where red color means that the left class has stronger support for this cluster and blue means that the right class has stronger support.}
\label{fig:bow_assignment_diff_exp3a_4vs5}
\end{center}
\end{figure}

We further investigate the performance on the GeoMat dataset. For GeoMat (s)PBoW significantly outperforms all other representations (see Table~\ref{tab:kernelResults}), first and foremost PI, which is the best related approach on this dataset in EXP-B. To this end, we generate confusion matrices for PI and PBoW that show that PBoW yields a better class separation (see Fig.~6 in SM$^a$).
Averaged PBoW histograms for two example classes (``cement smooth'' and ``concrete cast-in-place'') are shown in Fig.~\ref{fig:bow_assignment_diff_exp3a_4vs5}. For both classes the histograms are on the first sight similar (upper row in Fig.~\ref{fig:bow_assignment_diff_exp3a_4vs5}). However, by zooming-in towards the birth-persistence plane in Fig.~\ref{fig:bow_assignment_diff_exp3a_4vs5} (bottom row), differences become visible. The plots in the center illustrate the difference between the class distributions (red color means left class is stronger, blue means right class is stronger for this cluster). The classes distinguish themselves by fine-grained spatial differences. The set of three blue points around birth time of 0 (which are characteristic for class ``concrete cast-in-place'') surrounded by red points (which are characteristic for class ``cement smooth") illustrates this well (see lower central plot). For the discrimination of these two classes a particularly fine-grained codebook with many clusters is needed. The PI has problems with such fine-grained structures, because due to its limited resolution, all topological components in the most discriminative area would most likely fall into one pixel. Therefore, an extraordinary high resolution would be necessary to capture the discriminative patterns between those two classes. The bag-of-words model makes our approaches independent of the resolution and enables to efficiently capture such fine-grained differences. More examples from the GeoMat dataset can be found in SM$^a$ in Section 5.4 together with additional evaluations of (s)PBoW on computation time, dataset size and accuracy, see Sections~5.2 and 5.3.

% ======================================================================
% ======================================================================
%\section{Conclusion}
%\label{sec:conclusion}
%We have introduced persistence bag-of-words, a novel vectorial representation for PDs, which has attractive theoretic properties, i.e. stability with respect to 1-Wasserstein distance as well as practical properties, i.e compactness, expressiveness, and spatial adaptivity (independence on resolution). Experiments on six heterogeneous datasets show that it achieves state-of-the-art and partly even better performance while requiring significantly less computation time. The high computational efficiency could in future facilitate the application of TDA to larger datasets than possible today. Ongoing and future research focuses on extending persistence BoW to more comprehensive BoW encodings, such as VLAD~\cite{jegou2010aggregating} and Fisher Vectors~\cite{perronnin2007fisher}.

% ======================================================================
% ======================================================================
\section*{Acknowledgements}
This work was supported by the National Science Centre, Poland under grants no. 2015/19/D/ST6/01215, 2015/19/B/ST6/01819 and 2017/25/B/ST6/01271, the Austrian Research Promotion Agency FFG under grant no. 856333 and 866855, the Lower Austrian Research and Education Company NFB under grant no. LSC14-005 and by EPSRC grant no. EP/R018472/1.

{\small
\bibliographystyle{named}
\bibliography{ijcai19}
}

\end{document}

\maketitle

% ======================================================================
% ======================================================================
\section{Introduction}

This document represents the supplementary material to the paper ``Persistence Bag-of-Words for Topological Data Analysis". Here we present additional background information as well as complementary information on the experimental setup and additional results of our analysis that could not be included in the paper due to spatial limitations.

% ======================================================================
% ======================================================================
\section{Background on Persistent Homology}
\label{persistence}

In this section we present basic introduction to persistent homology. Please consult ~\cite{edelsbrunner2010computational,EdLeZo2002,zomorodian2005computing} for more information. 

Topological spaces are typically infinite objects and for the sake of data analysis they have to be finitely represented by simplified objects called \emph{cell complexes}. Cell complexes are build from \emph{cells}: topologically simple objects having the property that an intersection of every pair of cells is either empty, or yet another cell in the cell complex.

A \emph{simplicial complex} is a particular instance of a general cell complex. It is a natural tool in the study of multi-dimensional point cloud data. Cells of simplicial complex are called \emph{simplices} and, in this particular case, are formed with convex hulls of collections of nearby points in the point cloud. Simplices are uniquely characterized by collection of points involved in their convex hulls.  

Given a point cloud $X$ with a distance $d$ and a parameter $r > 0$ one can define a {\em Vietoris-Rips complex} $VR(X,r)$. It is a simplicial complex whose every simpliex $\sigma = \{ v_0, v_1, \ldots, v_k \}$ satisfy $d(v_i, v_j) \le r$ for every $i,j \in \{ 0,\ldots,k \}$. 
For every simplex $\sigma \in VR(X,r)$ one can define a diameter of $\sigma$ being the largest distance between the points in $\sigma$. This gives a natural ordering of simplices in $VR(X,r)$: primarily by diameter of simplices and secondarily (when diameters of two simplices are the same) by inverse of the number of points in simplices\footnote{A number of points involved in the simplex minus one is a \emph{dimension} of the simplex.}. It is easy to see that every prefix of such a ordering forms a simplicial complex, and therefore any increasing sequence of numbers $0 < r_1 < r_2 < \ldots < r_n$ yields a nested sequence of simplicial complexes:
\begin{align*}
\emptyset \subset X=VR(X,0)\subset VR(X,r_1) \subset \\     VR(X,r_2) \subset \cdots\subset VR(X,r_n)
\end{align*}

Another typical scenario when such a nested sequences of cell complexes arises is the case of values of a function $f$ discretized on a grid $G$. The function $f : G \rightarrow \mathbb{R}$ is typically an output of some numerical method. The grid $G$ naturally corresponds to cubical complex $\mathcal{G}$, and the function $f$ provides an ordering of maximal cubes in the complex. This ordering induce a nested sequence of cubical complex, very much like a nested sequence of Vietoris-Rips complexes discussed above. 

To cover those and other possible cases later in this section we will focus on a general case of filtered cell complex:
\[
\emptyset = \mathcal{C}_0 \subset \mathcal{C}_1 \subset \cdots \subset \mathcal{C}_n = \mathcal{C}
\]
keeping in mind that most typically it will coming from a point cloud, or numerical simulations on a grid. 

Given each cell complex $\mathcal{C}_i$ in the filtration one can define its homology, $H(\mathcal{C}_i)$. Rather than give a formal definition which can be found in~\cite{edelsbrunner2010computational}, we will focus on the intuitive understanding of the concept. Homology in dimension $0$ measures number of connected components. In dimension $1$ it measures the cycles which do not bound a (deformed) surface. In dimension $2$ it corresponds to voids, i.e. regions of space totally bounded by a collection of triangles (very much like a ball bounds the void inside it). The idea of a cycle bounding a hole in the complex can be formalized using homology theory for arbitrary dimension. 

\begin{figure}[ht]
\centering
\includegraphics[width=0.98\linewidth]{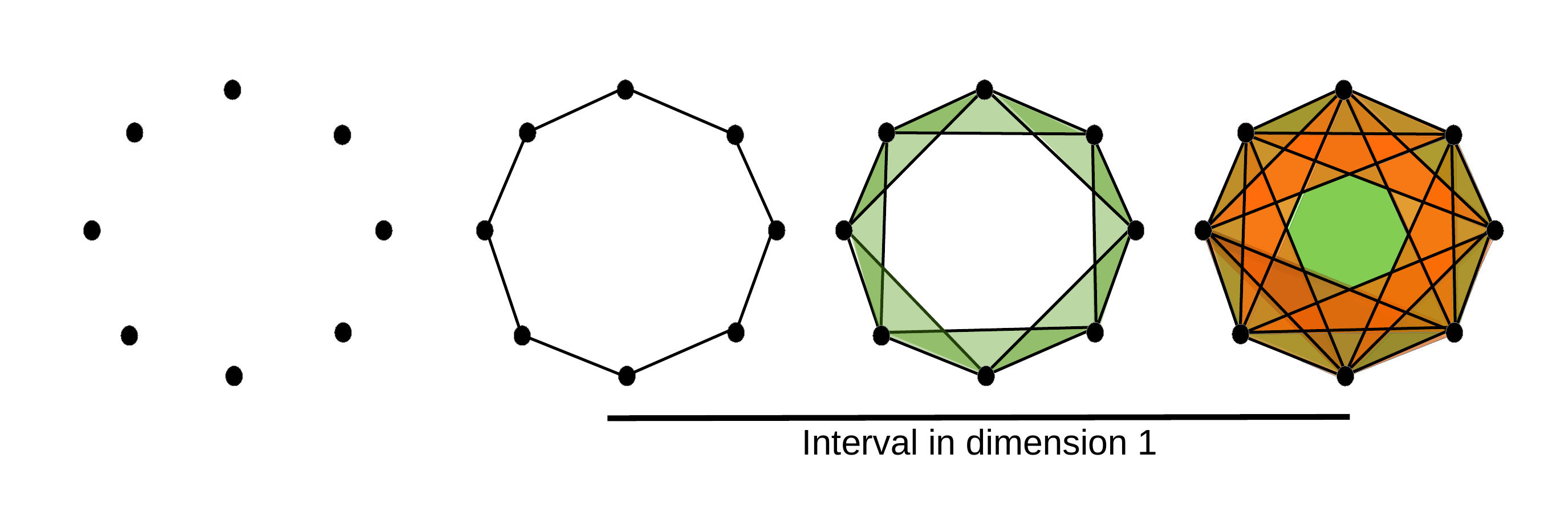}
\caption{Various stages of construction of a Vietoris-Rips complex for eight points sampled from a circle. Initially, for sufficiently small radius, only vertices are present in the complex. Gradually more and more edges along with higher dimensional simplices of increasing diameter are added. In all but the initial and final stage of the construction the topology of a circle is visible, and therefore will be recovered by PH in dimension one (depicted by the long bar below the picture).}
\label{fig:persistence_llustration}
\end{figure}

Persistent homology measures the evolution of homology for the constitutive complexes in a filtration. Once more and more cells are being added to a complex $\mathcal{C}_i$, new connected components or cycles may appear, old one may became trivial, or became identical (homologous) to others created earlier. For every connected component or a cycle, standard Persistent Homology captures the following two events:
\begin{itemize}
\item the first moment $b$, refereed as a \emph{birth time}, when it appears in the filtration,
\item the last moment $d$, referred to as \emph{death time}, when it either became trivial, or became identical to other cycle that has been created earlier.
\end{itemize}
Instead of a standard birth-death summaries of persistent homology in this paper we use birth-persistence coordinates, which can be obtained by the $[b,d] \rightarrow [b,d-b]$ transformation. 
It is useful to relate the abstract machinery of PH with geometric intuition. To do that let us concentrate on a Fig.~\ref{fig:persistence_llustration}. Over there, on the left, a collection of points sampled from a circle is presented. Moving to the right reveals various stages of construction of a Vietoris-Rips complex for an increasing sequence of radii. Along with increasing radius, simpices of increasing diameters are added to the complex. The key observation is that we can see a cycle for all the stages of the construction except from the initial and the final one. PH will capture this observation by outputting a long persistence interval in dimension one (visualized in the picture), spreading from the radius corresponding to the second to the radius corresponding to the last stage of the construction. This simple example illustrates the basic geometrical idea behind PH.

We have a couple of assumptions about PDs. Firstly, as our aim is to perform computations, we assume that persistence diagrams consist of finitely many points of nonzero persistence. Secondly, PDs may also contain infinite intervals that correspond to so called \emph{essential classes}, i.e. the cycles that are born but never dies. Those infinite intervals needs to be processed prior to the computations. There are at least three strategies one can apply:
\begin{enumerate}
\item To ignore infinite intervals, and use only the finite ones in the consideration. 
\item To substitute $+\infty$ in the death coordinates of the essential classes with a number $N$ chosen by the user. A logical choice would be a number which is larger than a filtration value of any cell in the considered complex. 
\item To build a pair of descriptors: one for finite, and one for infinite intervals and use them together as a final descriptor. 
\end{enumerate}
Given the available options in the numerical experiments presented in this paper we have chosen the simplest option, i.e. to ignore the infinite intervals. 
There are various classical metrics used to compare persistence diagrams~\cite{edelsbrunner2010computational}. We will review them here, as they are essential in the study of stability of the presented representatons. Note that the presentation is a bit non standard, as we are working on birth-peristence coordinates. Given two diagrams $B$ and $B'$, we construct a matching $\eta : B \rightarrow B'$ assuming that points can also be matched to $y=0$ axis. Putting $B$ and $B'$ in the same diagram, one can visualize a matching $\eta$ by drawing a line segment between $x \in B$ and $\eta(x)$ (note that $\eta(x)$ is either in $B'$, or is a projection of $x$ to its first coordinate). Given all the line segments, for each matching we can store the longest one, or a sum of lengths of all of them (to a power $q$). Taking the minimum over all possible matching of the obtained numbers will yield the \emph{bottleneck} distance in the first case, and the \emph{Wasserstein} distance (to the power $q$) in the second case. More formally we introduce following definition:

\medskip
\noindent
{\bf Definition} {\it q-Wasserstein distance} between two persistence diagrams $B,B'\in\D$ is defined as:
\[
W_q(B,B') := \left[\inf_{\eta: B\rightarrow B'} \sum_{x\in B} \norm{x-\eta(x)}^q_\infty\right]^\frac{1}{q}.
\]
In particular:
\[
W_1(B,B') := \inf_{\eta: B\rightarrow B'} \sum_{x\in B} \norm{x-\eta(x)}_\infty.
\]

An important feature of persistent homology is its \emph{stability}. Intuitively, it indicate that small changes in the filtration implies small changes (for instance in Wasserstein metric) in the resulting persistence diagrams. Formally we have following theorem:\medskip

\noindent {\bf Theorem} \cite{edelsbrunner2010computational} Let $X$ be a finite cell complex and $f,g : \mathbb{X} \rightarrow \mathbb{R}$ filtering Lipshitz functions. Let $B$ and $B'$ be the PDs of $\mathcal{X}$ with filtration induced by $f$ and $g$ respectively. Then there exist constants $C$ and $k$ such that $W_1(B,B') \leq C || f-g ||_{\infty}^{1-k}$.

In this paper, we show stability of sPBoW with respect to 1-Wasserstein distance. Combined with the stability result from above, this indicate stability of sPBoW with respect to the perturbation of initial data. 

% ======================================================================
% ======================================================================
\section{Complexity analysis}
\label{sec:complexity}

In our code, we apply VLFeat library~\cite{vedaldi08vlfeat}, both for generating codebook and histogram assignment. Codebook generation in PBoW consists of two steps: running heuristic k-means clustering and building a static k-d tree. Fast querying of such a tree is used in histogram assignment step. The complexity of the employed k-means implementation, which uses Lloyd's algorithm equals $O(N|D|I)$, where: $|D|$ is a cardinality of consolidated diagram, $N$ is a number of clusters, and $I$ is a number of iterations needed until convergence. The complexity of building the k-d tree equals $O(N|D|log(|D|))$. Therefore overall complexity is $O(N|D|(I+log(|D|)))$. Querying a single point in a balanced k-d tree with randomly distributed points takes $O(log(|D|))$ time on average. The codebook of sPBoW is generated with Gaussian mixture models, which implements the expectation maximization (EM) algorithm from~\cite{dempster1977maximum}. The complexity of EM varies strongly depending on implementation, but we could not find the information about the complexity of EM in library. However, according to our experiments, it works similarly fast as k-means clustering.

% ======================================================================
% ======================================================================
\section{Datasets and Experimental Setup}
\label{sec:datasets}

For our evaluation we incorporate a number of datasets which cover a wide range of different retrieval tasks. Firstly, to provide a proof-of-concept we evaluate all the approaches on a synthetically generated shape classes from~\cite{adams2017persistence}. Next, the approaches are evaluated on real-world datasets for geometry-informed material recognition (GeoMat)~\cite{degol2016geometry}, classification of social network graphs (Reddit)~\cite{hofer2017deep}, analysis of 3D surface texture (PetroSurf3D)~\cite{zeppelzauer2017study}, and 3D shape segmentation~\cite{Carrire2017SlicedWK}. All the datasets are discussed in details in the following sections.

As the computation times for some of the considered methods, especially for kernel based approaches, do not scale well with the sizes of datasets, we have decided to sub-sample the largest datasets in EXP-A by randomly selecting 30 (GeoMat), 100 (reddit-5k), 50 (reddit-12k) and 390 (PetroSurf3D) samples for each class.

For all datasets, except Reddit,  we consider the PDs of dimension 1 as a common input (cycles) since they best express the internal structure in the data and yielded the most promising results in related works~\cite{adams2017persistence,carriere2015stable}. In case of Reddit database we use PDs of dimension 0 (connected components), since we consider graphs as $1$-complex, thus first dimensional homology generators never die. In the considered datasets no infinite intervals of dimension 1 occur. In cases where infinite intervals are present, there are different ways to proceed: (1) ignoring them, (2) substituting infinity with some (large) number or (3) building separate representations for finite and infinite diagrams. In the general case, we recommend to compute persistence bag-of-words for PDs of all available dimensions separately and to combine them during modeling.
 
% ==============================
\subsection{Synthetic Dataset}

\begin{figure*}[ht]
\begin{center}
\includegraphics[width=0.7\linewidth]{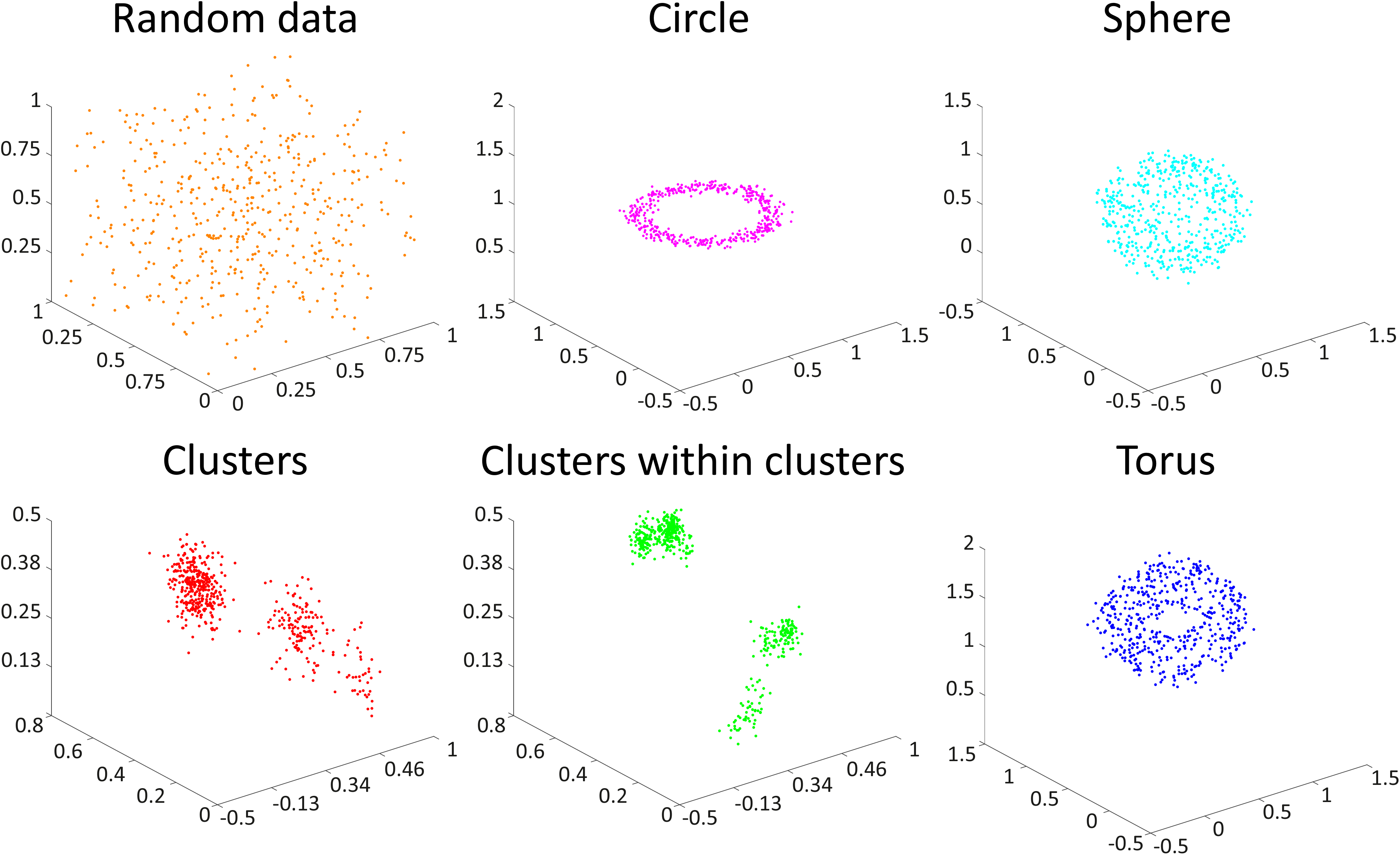}
\caption{Example shapes from the six shape classes of the synthetic dataset.}
\label{fig:syntheticData}
\end{center}
\end{figure*}

The first dataset we are going to consider is the synthetic dataset introduced by Adams et al.~\cite{adams2017persistence}. It consists of six shape classes represented by point clouds of the following geometrical objects: unit cube, circle of diameter one, sphere of diameter one, three clusters with centers randomly chosen from unit cube, hierarchical structure of three clusters within three clusters (where the centers of the minor clusters are chosen as small perturbations from the major cluster centers), and a torus (see Fig.~\ref{fig:syntheticData} for example shapes). Each point cloud is perturbed by positioning a Gaussian distribution of standard deviation $0.1$ at this point and sampling novel points from the distribution. Overall this dataset contains $50$ point clouds for each of the six classes, each containing $500$ 3D points. This gives $300$ point clouds in total. We subsample 80\% of the samples as training data and use the remaining samples for testing.

From each point cloud we compute the PDs in dimension $1$ for a Vietoris-Rips filtration for a radius parameter equal to the maximal distance between points in the point cloud. For this purpose, we employ the approximation method proposed by Dey et al.~\cite{Dey:2019:SET:3310279.3284360} and the SimBa implementation based on the work of Dayu Shi\footnote{\url{http://web.cse.ohio-state.edu/~dey.8/SimBa/Simba.html}, last visited October, 2018}.

% ==============================
\subsection{Geometry-Informed Material Recognition Dataset (GeoMat)}

The GeoMat dataset contains a collection of materials and provides geometry information (point clouds) as well as visual images of 19 different materials, such as ``brick", ``grass" and ``gravel"~\cite{degol2016geometry}. The GeoMat dataset contains patches sampled from larger photographs of surfaces from buildings and grounds. Each patch predominantly represents only one material. 
Each class consists of $600$ images, each of size $100\times 100$ pixels. Among them, there are pictures of different scales, i.e. $100\times 100$, $200\times 200$, $400\times 400$ and $800\times 800$. 

To keep dataset at a reasonable size for our experiments, we selected only patches of resolution $400\times 400$, which results in $149$ patches for each of the $19$ categories. For each patch, the dataset provides a depth image\footnote{Source: \url{http://web.engr.illinois.edu/~degol2/pages/MatRec_CVPR16.html}, last visited August 2018}. The depth images contain the local (fine-grained) surface texture as well as the global surface curvature. To filter out the global curvature we transformed each depth image into a point cloud in 3D space, consisting of $10000$ points (every point represents one of the $100\times 100$ pixels). Next, point cloud were rotated in a way that the Z axis represents depth, then removed the global surface curvature by fitting a second degree function (paraboloid) to the point cloud, which values were finally subtracted from the Z coordinates of the points. The values of the Z-coordinates were then centered at 0. Finally, the point cloud was projected back into a bitmap (depth map), with the Z-coordinates as depth values. Ultimately, we computed PDs by grayscale filtration. We employ the original train/test partition with 400/200 samples per class in our experiments. 

% ==============================
\subsection{Social Network Graphs Datasets (Reddit)}

To evaluate our approaches on a broad range of possible input data, we further incorporate graph-based datasets in our evaluation. To this end we employ the reddit-5k and reddit-12k datasets from~\cite{yanardag2015deep}, which contain discussion graphs from the reddit platform\footnote{Reddit is a content-aggregation webseite: \url{http:\\reddit.com}}. Nodes in the graphs correspond to users and edges between users exist if one user has commented a posting of the other user. Different graphs are labeled by subreddits, which refer to different topics. The dataset reddit-5k contains overall 5000 graphs for 5 popular subreddits. For the larger dataset reddit-12k we sample 5.643 graphs for 11 subreddits including topics like, e.g. ``worldnews", ``videos" and ``atheism". The task for both datasets is to predict the subreddit (topic) from the input graph. For both datasets we use the pre-computed PDs from\footnote{Source: \url{https://github.com/c-hofer/nips2017}, last visited October, 2018}. For Reddit experiments we employ the original ratio of 90\% graphs in the training set and the remaining 10\% in the test set.

% ==============================
\subsection{3D Surface Texture Dataset (PetroSurf3D)}

A further dataset in our experiments is the recently released \emph{PetroSurf3D} dataset, which contains high-resolution 3D surface reconstructions from the archaeological domain with a resolution of approximately 0.1mm~\cite{PetroSurf3D}. 
The reconstructions represent 26 natural rock surfaces that exhibit human-made engravings (so-called rock art) and thereby exhibit complex 3D textures. 
The classification task for PetroSurf3D is to automatically predict which areas of the surface contain engravings and which not, i.e. there are two classes of surface topographies: engraved areas and the natural rock surface. Engraved areas represent approximately 19\% of the data. For each surface a precise pixel-accurate ground truth exists together with a depth map of the surface. The depth maps are analyzed in a patch-wise manner. 
Overall, there are 754.386 patches to classify for all 26 surfaces. For our experiments we randomly select 13 surfaces as a training set and employ the remaining ones for evaluation. 
The training set is further sub-sampled (to keep the number of training samples in a practical range). Overall, a balanced set (equal class cardinalities) of 600 patches per surface ($13*600=7800$ samples) is used for training. The testing set contains a balanced subset of 600 random samples from each of the remaining 13 surfaces. For each patch a PD is computed by grayscale filtration as a basis for our experiments.

% ==============================
\subsection{3D Shape Segmentation Dataset}

We further employ the 3D shape dataset from~\cite{chen2009benchmark} which was preprocessed by Carri{\`e}re et al.~\cite{carriere2015stable} for topological data analysis. The preprocessed dataset contains PDs for $5700$ 3D points from airplane models. Each point is assigned to one sub-part (segment) of an airplane, e.g., 'wing', 'vertical stabilizer' and 'horizontal stabilizer'. For our experiments we use the PDs computed by Carri{\`e}re et al.~\cite{carriere2015stable} from their repository\footnote{Source: \url{https://github.com/MathieuCarriere/sklearn_tda}, last visited October 2018}. The PDs were generated by tracking topology evolution of a geodesic ball centered at the individual points of the input 3D model. The radius thereby grows from 0 to infinity. We focus on PDs of dimension $1$ as the considered 3D shapes are connected. 
The task is to classify each point according to the segment it belongs to. By classifying each point a segmentation of the airplane into its parts is finally obtained. In our experiments we employ the original 50/50 train-test split.

% ======================================================================
% ======================================================================
\section{Evaluated Parameters}

To find optimal parameters for each evaluated approach, we run a grid search over their respective hyperparameters. For each parameter combination we run a complete classification experiment including cross-validation to evaluate its performance. We repeat each experiment 5 times with  randomly selected training and test partitions and average the achieved performances over all 5 repetitions. The only exception is PetroSurf3D where we divided the set of all surfaces into 4 folds (resulting in four repetitions) according to original work in \cite{PetroSurf3D}.  The hyperparameters and their evaluated values for each approach are listed in Table~\ref{tab:kernelParameters}, both for EXP-A and EXP-B. The optimal parameters are highlighted in bold.

\onecolumn
\begin{sidewaystable}
\begin{scriptsize}
\centering
\begin{tabular}{l|c|c|c|c|c|c|c}
\hline
\toprule
\multicolumn{8}{c}{\thead{{\bf EXP-A}}} \\
\textbf{Descr.} & {} & 
    \textbf{Synthetic} & 
    \textbf{GeoMat} & 
    \textbf{Reddit-5k} & 
    \textbf{Reddit-12k} & 
    \textbf{PetroSurf3D} & 
    \textbf{3D Shape Segm.}\\
\hline
\textbf{MK} & $\sigma$ & 
    \thead{$\{\textbf{0.5}, 1, 2\}$} &
    \thead{$\{0.5, \textbf{1}, 2\}$} &
    \thead{$\{\textbf{0.5}, 1, 2\}$} &
    \thead{$\{0.5, 1, \textbf{2}\}$} &
    \thead{$\{\textbf{0.5}, 1, 2\}$} &
    \thead{$\{\textbf{0.5}, 1, 2\}$} \\
\hline
\textbf{SWK} & $n$ & 
    \thead{$\{\textbf{50}, 100, 150, 200, 250\}$} &
    \thead{$\{\textbf{50}, 100, 150, 200, 250\}$} &
    \thead{$\{\textbf{50}, 100, 150, 200, 250\}$} &
    \thead{$\{\textbf{50}, 100, 150, 200, 250\}$} &
    \thead{$\{\textbf{50}, 100, 150, 200, 250\}$} &
    \thead{$\{\textbf{50}, 100, 150, 200, 250\}$} \\
\hline
% EXP-A params PI
\multirow{5}{*}{\textbf{PI}} & $r$ & 
        \thead{$\{10,20,...,\textbf{50},...100\}$} &
        \thead{$\{10,20,40,60,80,\textbf{100}\}$} &
        \thead{$\{10,20,30,40,50,$\\$60,80,100,\textbf{120}\}$} &
        \thead{$\{10,20,30,\textbf{40},50,$\\$60,80,100,120\}$} &
        \thead{$\{10,20,...\textbf{80},...100\}$} &
        \thead{$\{10,20,...\textbf{70},...100\}$}
        \\
    {}  & $\sigma$ & 
        \thead{$\{\textbf{0.1}, 0.5, 1, 2\}$} & 
        \thead{$\{\textbf{0.5}, 1, 2\}$} & 
        \thead{$\{\textbf{0.5}, 1, 2, 3\}$} & 
        \thead{$\{\textbf{0.5}, 1, 2, 3\}$} & 
        \thead{$\{0.5, \textbf{1}, 2\}$} & 
        \thead{$\{\textbf{0.5}, 1, 2, 3\}$} \\
    {}  & $\sfw$ & 
        $\{\textbf{\sfw}, \sfnw\}$ & 
        $\{\sfw, \textbf{\sfnw}\}$ &
        $\{\sfw, \textbf{\sfnw}\}$ & 
        $\{\sfw, \textbf{\sfnw}\}$ &
        $\{\textbf{\sfw}, \sfnw\}$ & 
        $\{\textbf{\sfw}, \sfnw\}$\\  
\hline
% EXP-A params RM
\multirow{5}{*}{\textbf{RM}} & $r$ &
        $\thead{\{\textbf{10}, 20, 40\}}$ & 
        $\thead{\{\textbf{10}, 20, 40, 60\}}$ & 
        $\thead{\{\textbf{10}, 20, 40, 60\}}$ & 
        $\thead{\{10, 20, \textbf{40}, 60\}}$ & 
        $\thead{\{10, 20, 40, \textbf{60}\}}$ & 
        $\thead{\{\textbf{10}, 20, 40, 60\}}$ \\
    {}  & $\sigma$ &
        $\{0.1, \textbf{0.2}, 0.3\}$ &
        $\{\textbf{0.1}, 0.2, 0.3\}$ &
        $\{\textbf{0.1}, 0.2, 0.3\}$ &
        $\{\textbf{0.1}, 0.2, 0.3\}$ &
        $\{\textbf{0.1}, 0.2, 0.3\}$ &
        $\{\textbf{0.1}, 0.2, 0.3\}$ \\
    {}  & $d$ & 
        $\thead{\{\textbf{50}, 75, 100\}}$ &
        $\thead{\{50, \textbf{75}, 100\}}$ &
        $\thead{\{50, \textbf{75}, 100\}}$ &
        $\thead{\{50, 75, \textbf{100}\}}$ &
        $\thead{\{50, \textbf{75}, 100\}}$ &
        $\thead{\{\textbf{50}, 75, 100\}}$ \\
\hline
% EXP-A params PBOW
\multirow{2}{*}{\textbf{PBoW}} & $N$ & 
        \thead{$\{10,20,30,...\textbf{200}\}$} & 
        \thead{$\{10,20,\textbf{30},40,60,$\\$80,...200\}$} & 
        \thead{$\{10,20,\textbf{30},...,60,$\\$80,100,...200\}$} & 
        \thead{$\{10,20,30,40,\textbf{50},$\\$60,80,100,120\}$} &
        \thead{$\{10,20,30,40,50,$\\$60,80,...\textbf{160},...200\}$} &
        \thead{$\{10,20,...,\textbf{80},...100\}$}
        \\
    {}  & $\sfw$ & 
        $\{\textbf{\sfw}, \sfnw\}$ & 
        $\{\sfw, \textbf{\sfnw}\}$ &
        $\{\sfw, \textbf{\sfnw}\}$ & 
        $\{\sfw, \textbf{\sfnw}\}$ &
        $\{\sfw, \textbf{\sfnw}\}$ & 
        $\{\textbf{\sfw}, \sfnw\}$\\ 
    {} & $S$ &
        \thead{$\{1000, 5000, \textbf{10000}\}$} &
        \thead{$\{\textbf{2000}, 10000, 50000\}$} &
        \thead{$\{\textbf{2000}, 10000, 50000\}$} &
        \thead{$\{5000, \textbf{10000}, 50000\}$} &
        \thead{$\{\textbf{5000}, 10000, 50000\}$} &
        \thead{$\{5000, \textbf{10000}, 20000\}$} \\
\hline
% EXP-A params sPBOW
\multirow{2}{*}{\textbf{sPBoW}} & $N$ & 
        \thead{$\{10,20,...\textbf{140},...200\}$} & 
        \thead{$\{10,20,\textbf{30},40,60,$\\$80,...200\}$} & 
        \thead{$\{10,\textbf{20},...,60,$\\$80,...200\}$} & 
        \thead{$\{\textbf{10},20,30,40,50,$\\$60,80,100,120\}$} &
        \thead{$\{10,20,30,40,50,$\\$60,80,\textbf{100}...200\}$} &
        \thead{$\{10,20,...\textbf{70},...100\}$}
        \\
    {}  & $\sfw$ &
        $\{\textbf{\sfw}, \sfnw\}$ & 
        $\{\textbf{\sfw}, \sfnw\}$ &
        $\{\sfw, \textbf{\sfnw}\}$ &
        $\{\textbf{\sfw}, \sfnw\}$ &
        $\{\textbf{\sfw}, \sfnw\}$ & 
        $\{\textbf{\sfw}, \sfnw\}$\\     
    {} & $S$ &
        \thead{$\{1000, \textbf{5000}, 10000\}$} &
        \thead{$\{\textbf{2000}, 10000, 50000\}$} &
        \thead{$\{2000, \textbf{10000}, 50000\}$} &
        \thead{$\{\textbf{5000}, 10000, 50000\}$} &
        \thead{$\{5000, 10000, \textbf{50000}\}$} &
        \thead{$\{5000, 10000, \textbf{20000}\}$} \\
\midrule
\multicolumn{8}{c}{\thead{{\bf EXP-B}}} \\
\textbf{Descr.} & {} & 
    \textbf{Synthetic} & 
    \textbf{GeoMat} & 
    \textbf{Reddit-5k} & 
    \textbf{Reddit-12k} & 
    \textbf{PetroSurf3D} & 
    \textbf{3D Shape Segm.}\\
\hline
% EXP-B params PI
\multirow{5}{*}{\textbf{PI}} & $r$ & 
    \multirow{5}{*}{same as EXP-A} & 
        \thead{$\{10,20,30,40,50,\textbf{60}\}$} &
        \thead{$\{10,20,30,40,50,\textbf{60}\}$} &
        \thead{$\{10,20,30,40,\textbf{50},60\}$} &
        \thead{$\{10,20,30,40,\textbf{50},60\}$} &
    \multirow{5}{*}{same as EXP-A} \\
    {} & $\sigma$ & 
    {} & 
        \thead{$\{\textbf{0.5}, 1, 2, 3\}$} & 
        \thead{$\{\textbf{0.5}, 1, 2\}$} & 
        \thead{$\{\textbf{0.5}, 1, 2\}$} & 
        \thead{$\{0.5, 1, \textbf{2}\}$}  &
    {} \\
    {} & $\sfw$ &
    {} & 
        $\{\sfw, \textbf{\sfnw}\}$ &
        $\{\sfw, \textbf{\sfnw}\}$ &
        $\{\sfw, \textbf{\sfnw}\}$ &
        $\{\textbf{\sfw}, \sfnw\}$ &
    {} \\
\hline
% EXP-B params RM
\multirow{5}{*}{\textbf{RM}} & $r$ &
    \multirow{5}{*}{same as EXP-A} &
        $\thead{\{\textbf{10}, 20, 40, 60\}}$ &
        $\thead{\{\textbf{20}, 40, 60\}}$ &
        $\thead{\{\textbf{20}, 40, 60\}}$ &
        $\thead{\{20, \textbf{40}, 60\}}$  &
    \multirow{5}{*}{same as EXP-A} \\
    {} & $\sigma$ &
    {} &
        $\{\textbf{0.1}, 0.2, 0.3\}$ &
        $\{\textbf{0.1}, 0.2, 0.3\}$ &
        $\{\textbf{0.1}, 0.2, 0.3\}$ &
        $\{0.1, 0.2, \textbf{0.3}\}$  &
    {} \\
    {} & $d$ &
    {} &
        $\thead{\{\textbf{50}, 75, 100\}}$ &
        $\thead{\{\textbf{25}, 50, 75, 100\}}$ &
        $\thead{\{\textbf{25}, 50, 75, 100\}}$ &
        $\thead{\{25, 50, \textbf{75}, 100\}}$ &
    {} \\
\hline
% EXP-B params PBOW
\multirow{2}{*}{\textbf{PBoW}} & $N$ & 
    \multirow{2}{*}{same as EXP-A} &
        \thead{$\{10,20,40,60,$\\$...\textbf{100},...200\}$} & 
        \thead{$\{10,20,...60,$\\$80,\textbf{100}...200\}$} & 
        \thead{$\{10,20,...50,$\\$60,80,\textbf{100}\}$} & 
        \thead{$\{10,20,...\textbf{50},$\\$60,80,100\}$} &
    \multirow{2}{*}{same as EXP-A} \\
    {} & $\sfw$ &
    {} &
    $\{\sfw, \textbf{\sfnw}\}$ &
    $\{\sfw, \textbf{\sfnw}\}$ &
    $\{\sfw, \textbf{\sfnw}\}$ &
    $\{\sfw, \textbf{\sfnw}\}$ &
    {} \\
    {} & $S$ &
        {} &
        \thead{$\{2000, \textbf{10000}, 50000\}$} &
        \thead{$\{\textbf{2000}, 10000, 50000\}$} &
        \thead{$\{5000, 10000, \textbf{50000}\}$} &
        \thead{$\{\textbf{5000}, 10000, 50000\}$} 
        {} \\
% EXP-B params sPBOW
\hline
\multirow{2}{*}{\textbf{sPBoW}} & $N$ & 
    \multirow{2}{*}{same as EXP-A} &
        \thead{$\{10,\textbf{20},40,60, ...200\}$} & 
        \thead{$\{10,20,\textbf{30}...60,$\\$80,...200\}$} & 
        \thead{$\{10,20,...50,$\\$60,80,\textbf{100}\}$} & 
        \thead{$\{10,20,...50,$\\$60,80,\textbf{100}\}$} &
    \multirow{2}{*}{same as EXP-A} \\
    {} & $\sfw$ &
    {} &
        $\{\textbf{\sfw}, \sfnw\}$ &
        $\{\sfw, \textbf{\sfnw}\}$ &
        $\{\sfw, \textbf{\sfnw}\}$ &
        $\{\textbf{\sfw}, \sfnw\}$ &
    {} \\
    {} & $S$ &
        {} &
        \thead{$\{\textbf{2000}, 10000, 50000\}$} &
        \thead{$\{2000, \textbf{10000}, 50000\}$} &
        \thead{$\{5000, \textbf{10000}, 50000\}$} &
        \thead{$\{5000, \textbf{10000}, 50000\}$} 
        {} \\
\bottomrule
\end{tabular}
\caption{Parameters tested for EXP-A and EXP-B. The optimal parameters are highlighted in bold. Abbreviations: $\sigma$ is scale parameter in MK, $n$ is the number of lines slicing the plane in SWK, $r$ is resolution of PI or density map in RM, $\sigma$ is the sigma of the Gaussians employed, $\sfw$ is weighting (either no weighting (\sfnw) or with weighting (\sfw), $d$ is the dimension of rincipal geodesic analysis on the hypersphere, $N$ is the number of codewords of persistence bag-of-words.}
\label{tab:kernelParameters}
\end{scriptsize}
\end{sidewaystable}
\twocolumn

% ======================================================================
% ======================================================================
\section{Additional Results}

In this section we provide additional investigations on selected aspects of the novel representations and provide additional illustrations and figures which were not included in paper due to spatial limitations.

% ==============================
\subsection{Accuracy vs. Codebook Size}

\begin{figure*}
\centering
\begin{tabular}{c c}
\includegraphics[width=0.45\linewidth]{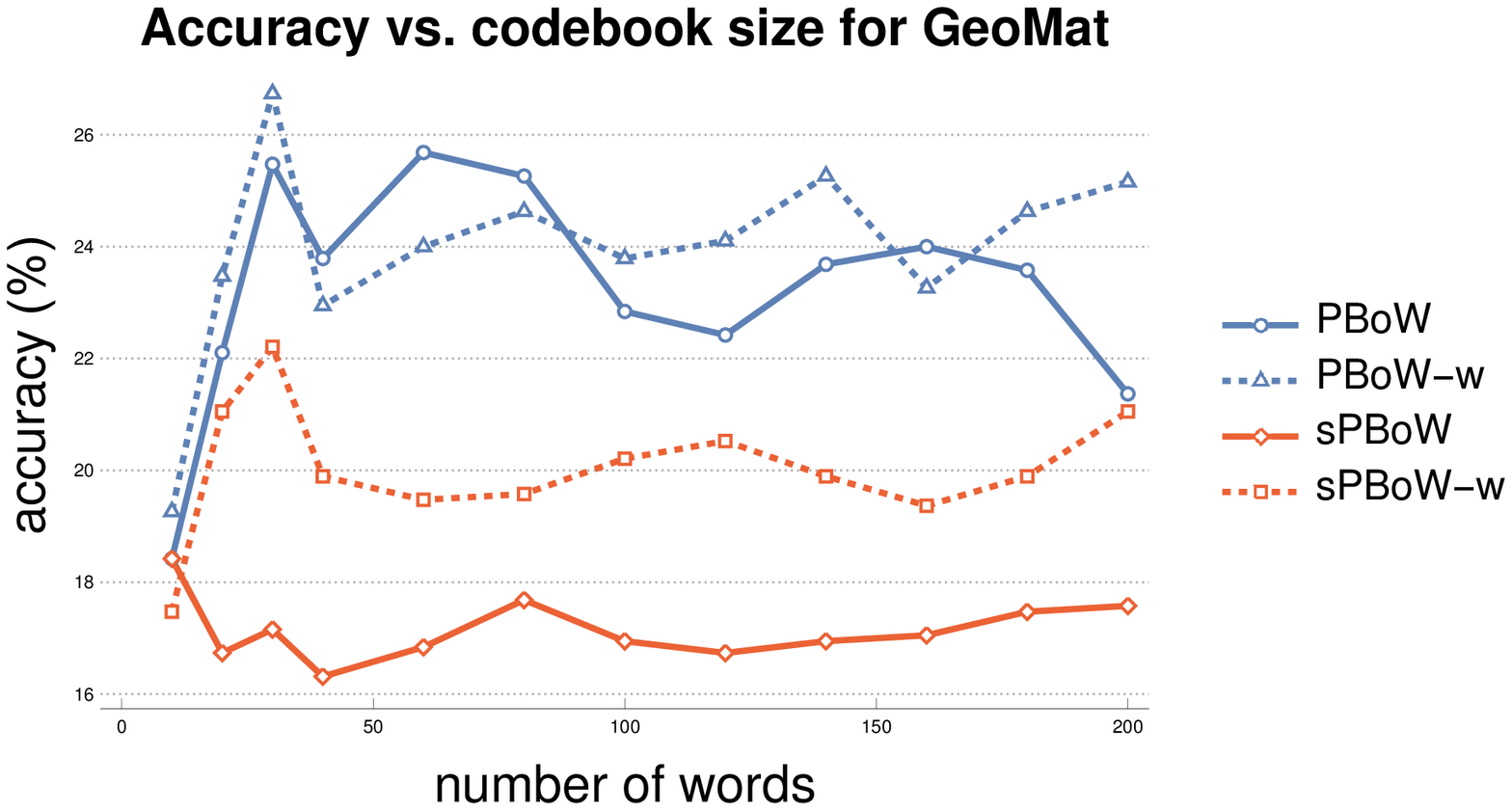} &
\includegraphics[width=0.45\linewidth]{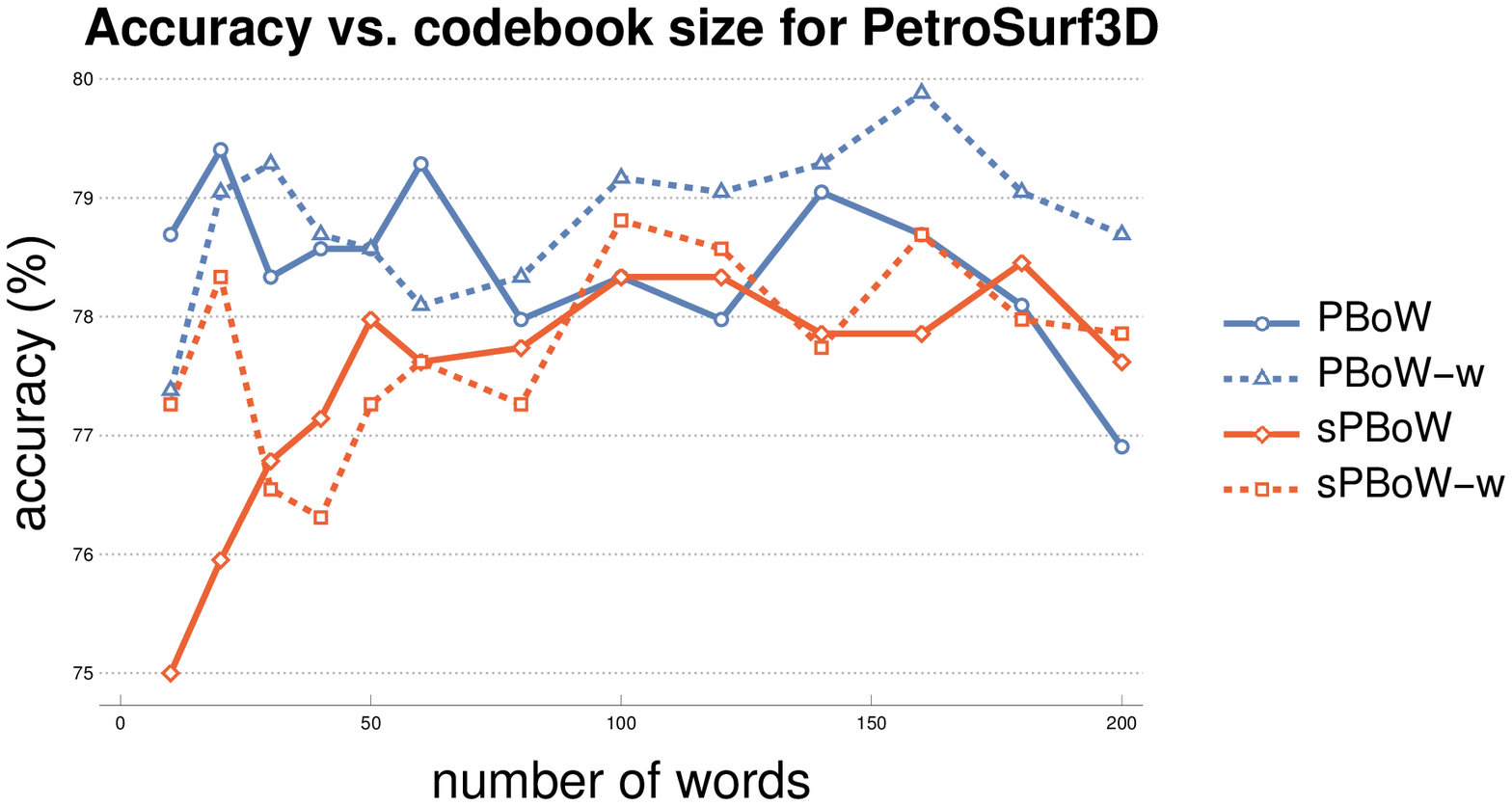} \\
\includegraphics[width=0.45\linewidth]{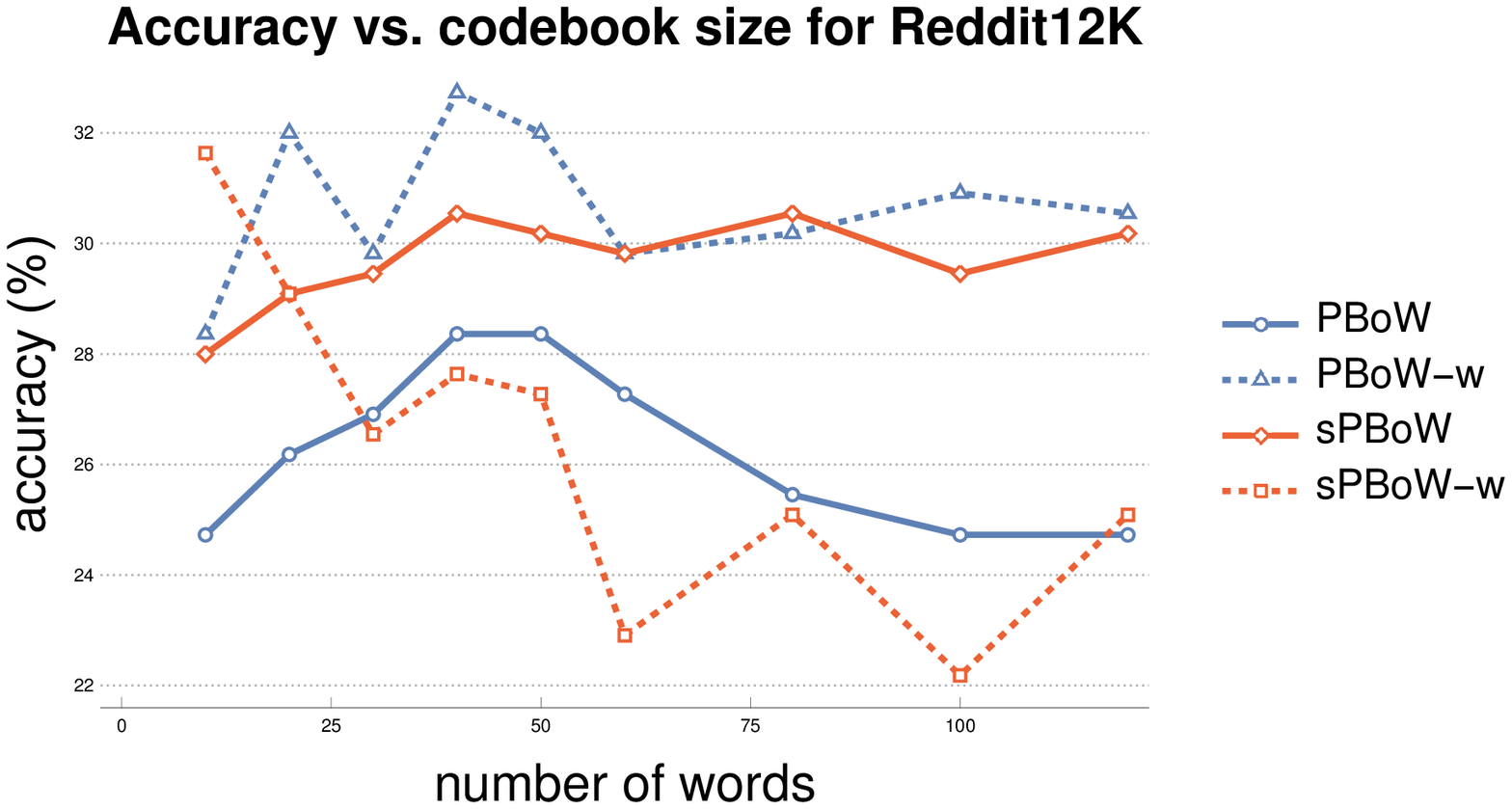} &
\includegraphics[width=0.45\linewidth]{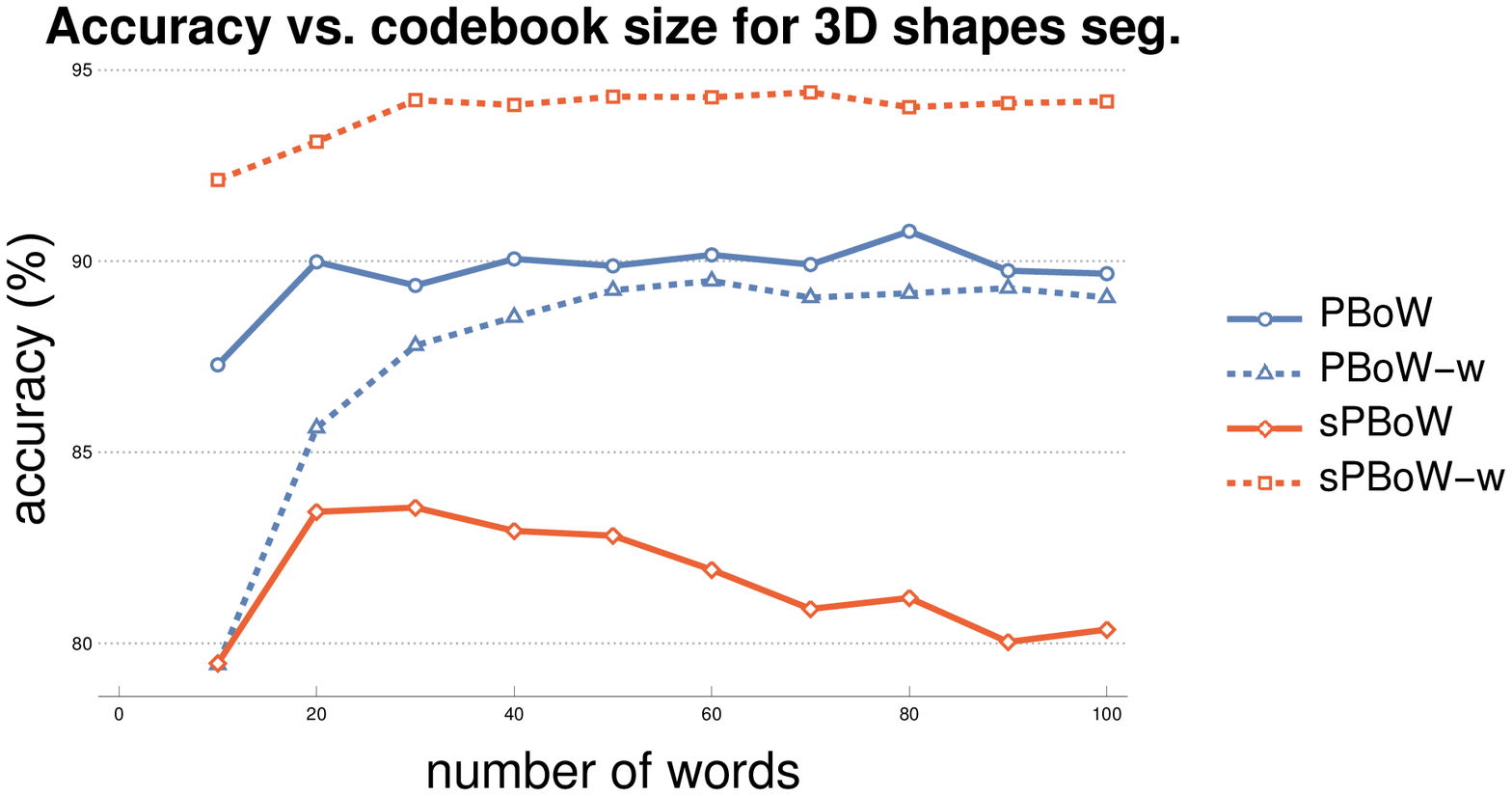}
\end{tabular}
\caption{Accuracy vs. size of a codebook for datasets from EXP-A without (solid lines) and with codebook weighting (dashed lines).}
\label{fig:acc_words_exp1}
\end{figure*}

In Fig.~\ref{fig:acc_words_exp1} we present the sensitivity of the proposed approaches on the codebook size for the remaining datasets to complement the paper.

% ==============================
\subsection{Time vs. Dataset Size}

\begin{figure}
\centering
\includegraphics[width=0.9\linewidth]{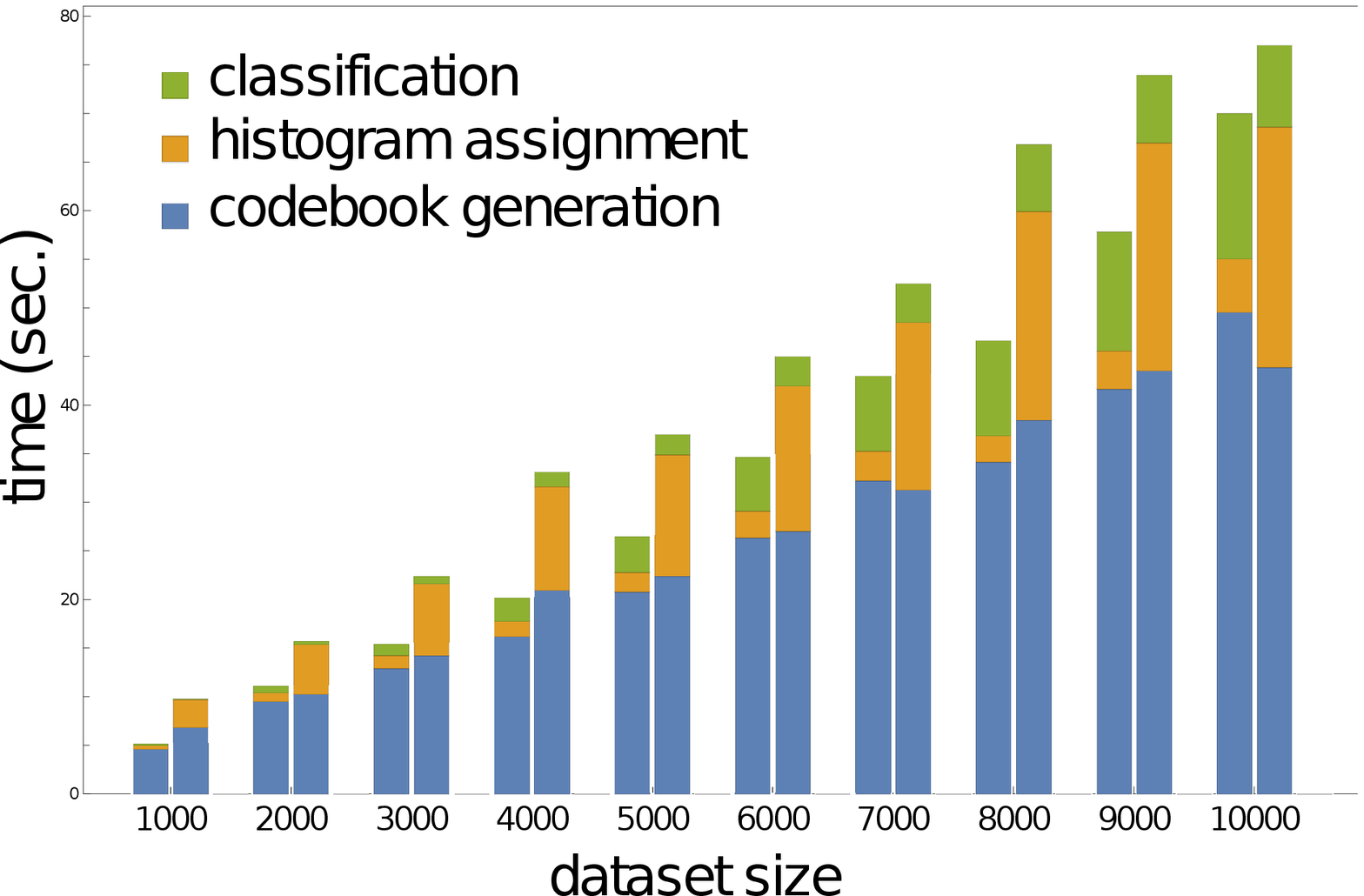}
\caption{Time vs. dataset size for PBoW and sPBoW (left and right bar of each pair), both with $N=50$. We measure the time needed for codebook generation (blue, bottom), histogram assignment (orange, middle), and classification (green, top), separately.}
\label{fig:time_datasize}
\end{figure}

To investigate the runtime behavior of the proposed approaches in more detail, we evaluate how they scale for a varying number of input PDs, i.e. dataset size. To this end, we adapt the largest dataset (PetroSurf3D) and randomly sample different numbers of PDs, i.e. from $1000$ to $10000$ in steps of $1000$. To get a detailed breakdown of computation time we measure the time needed for codebook generation, histogram assignment, and classification, separately. The computation of the PDs is not included in this breakdown. From the results presented in Fig.~\ref{fig:time_datasize}, we conclude that runtime grows linearly with the number of PDs. Most time is spent on codebook generation, where all PDs have to be consolidated and clustering is applied. Histogram assignment is faster in case of PBoW, because it is based on k-d trees~\cite{bentley1975multidimensional}, while in case of sPBoW, Gaussian likelihood has to be computed. One can also observe that SVM converge faster in case of the stable representation. The computational overhead of the sPBoW is in most cases small and grows only slowly with dataset size.

% ==============================
\subsection{Accuracy vs. Time}

Table~1 of the paper shows that our approaches achieve state-of-the-art or even better performance on almost all of the evaluated datasets. Furthermore they outperform all methods in speed. While the table shows only results for the optimal parameters (from the point of view of classification accuracy), here, we analyze the relation between accuracy and time in more detail. For this purpose, we use PI as a reference for comparison, as it represents the strongest competitor (in the sense of accuracy) of the proposed representations. 

\begin{figure*}
\centering
\begin{tabular}{c c}
\includegraphics[width=0.45\linewidth]{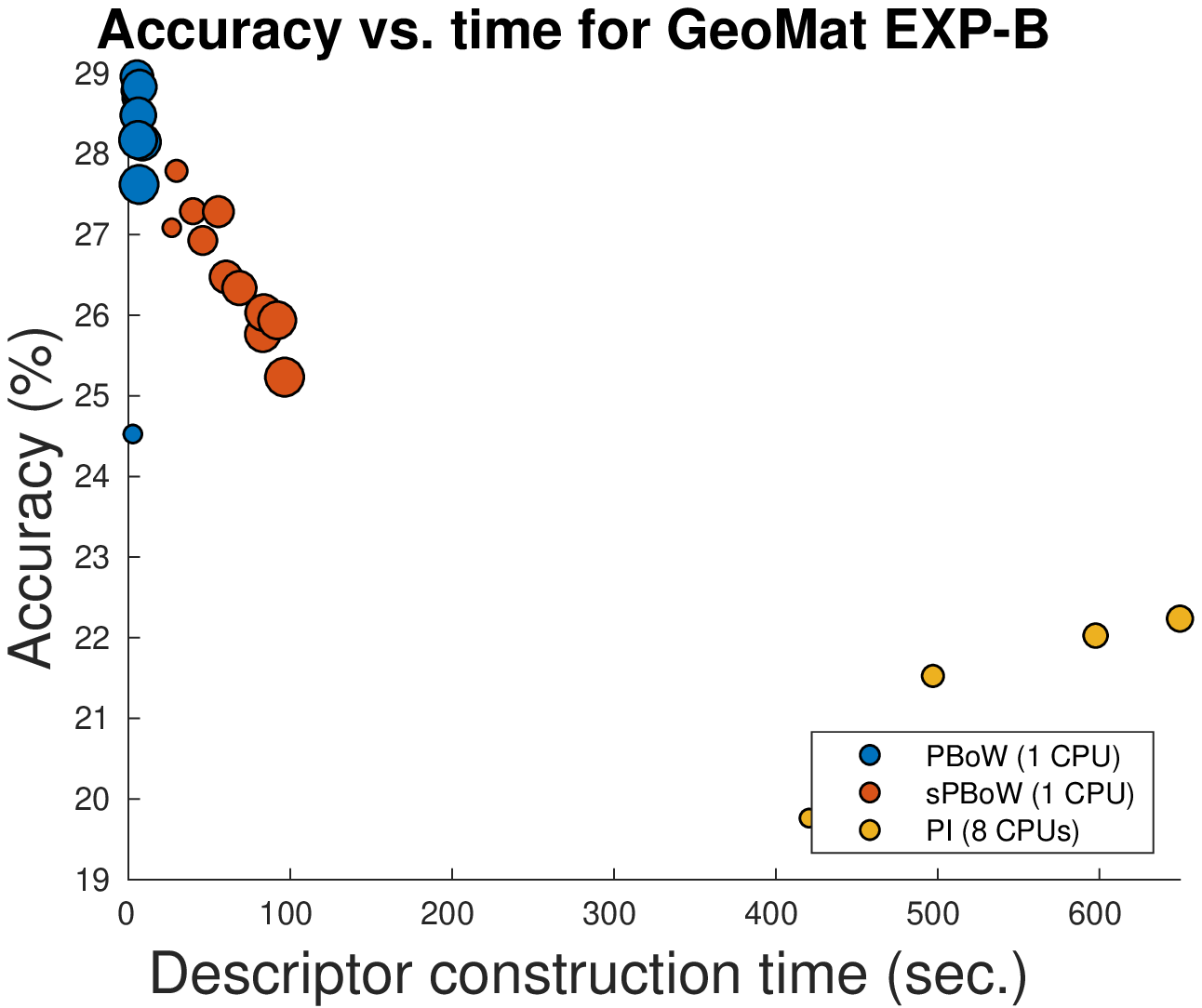} &
\includegraphics[width=0.45\linewidth]{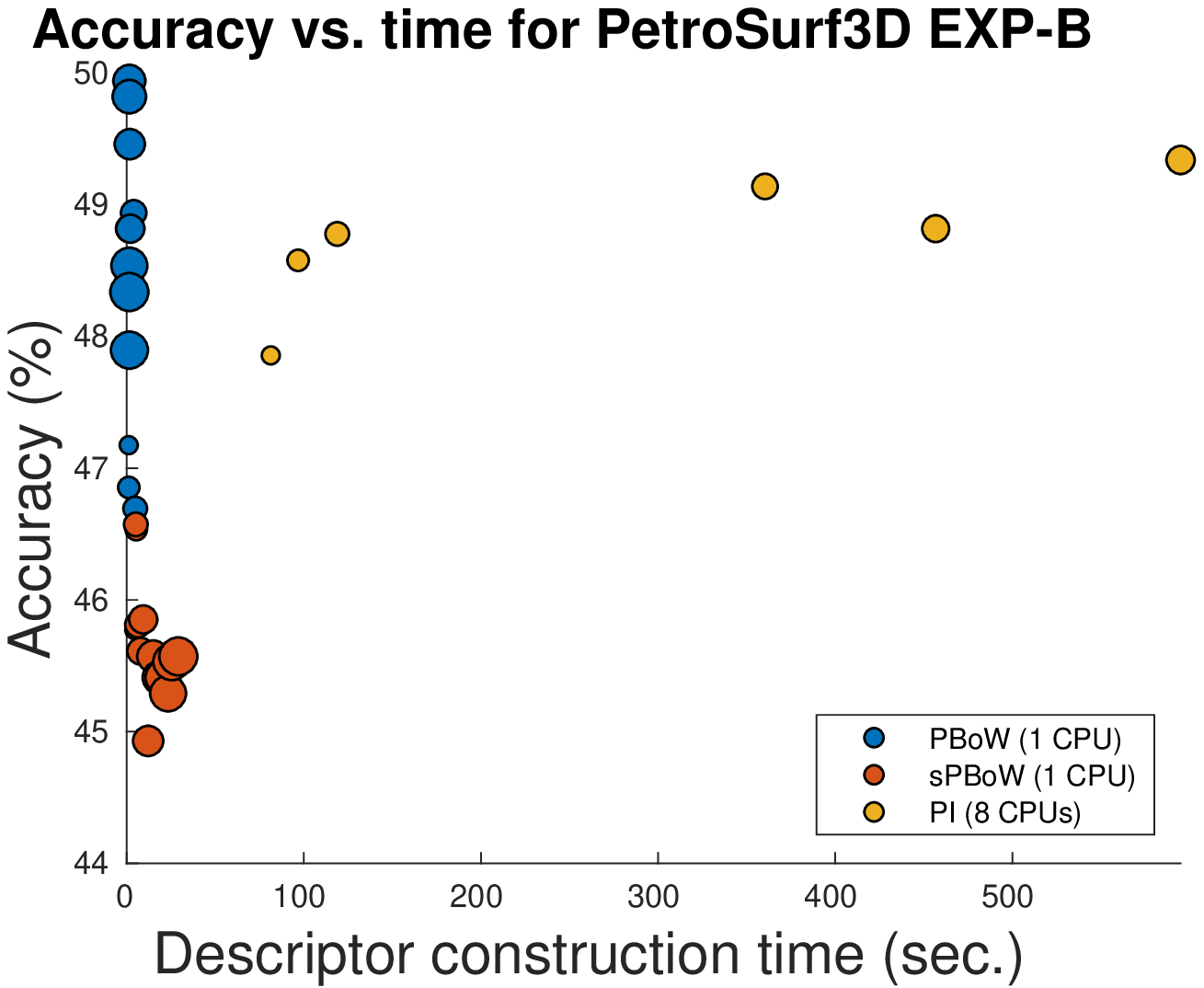} \\ \includegraphics[width=0.45\linewidth]{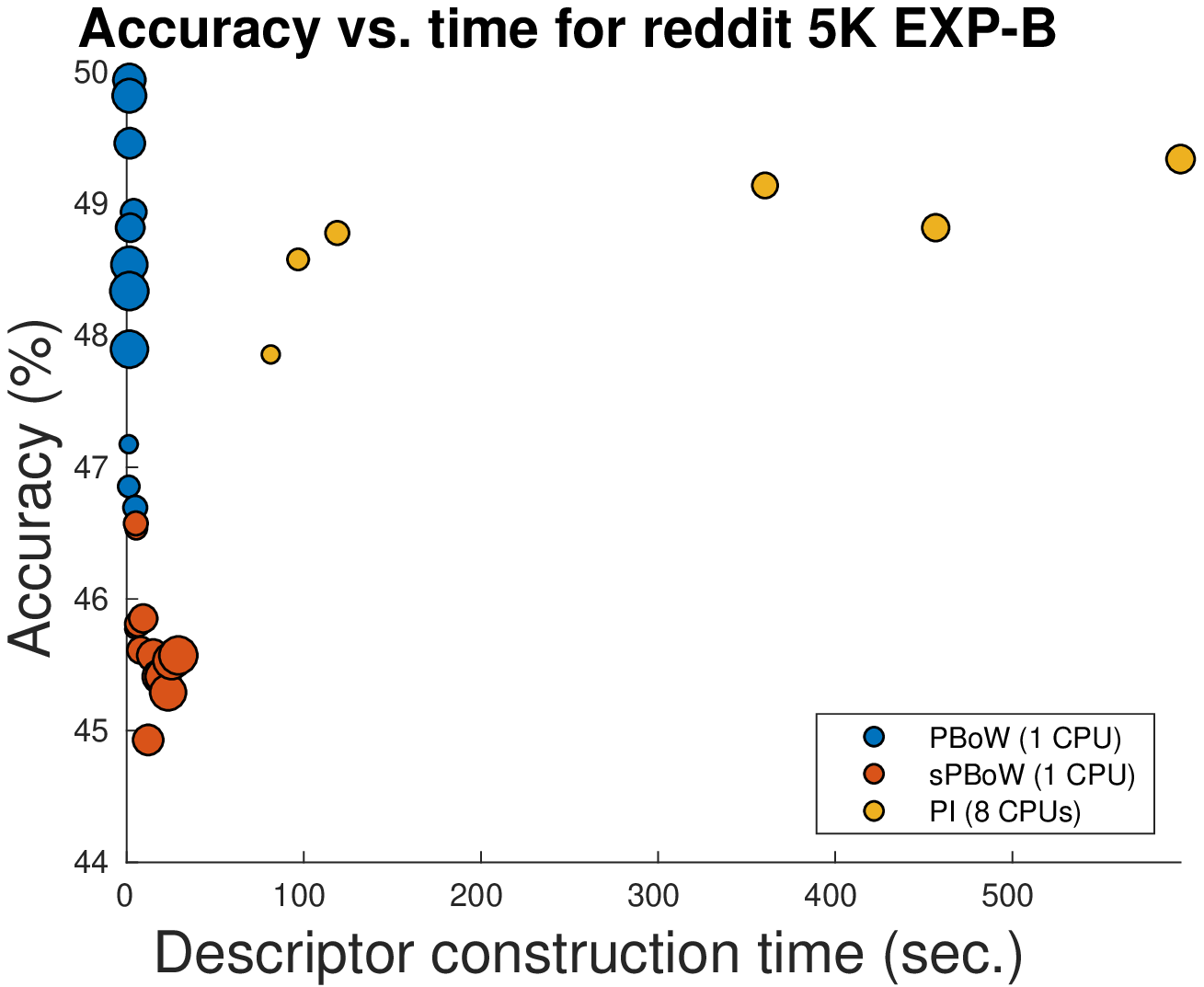} &
\includegraphics[width=0.45\linewidth]{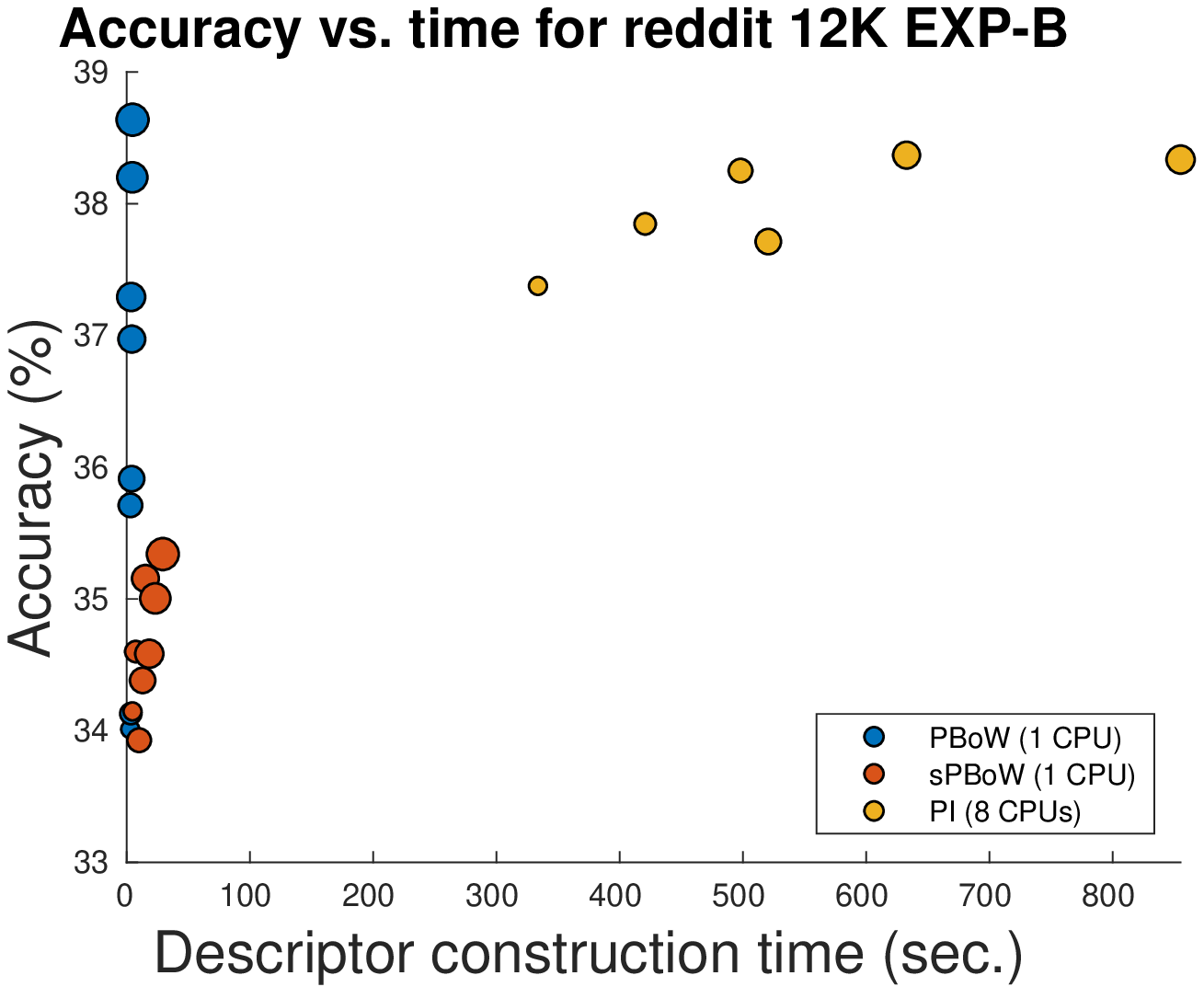}
\end{tabular}
\caption{Accuracy vs. time for PBoW and sPBoW compared with PI (the strongest related representation) applied to four datasets from EXP-B. The size of colored points represents the size $N$ of codebooks or the resolution $r$ of PI (evaluated values for $N$ and $r$ are those listed in Table~\ref{tab:kernelParameters} of the supplementary material). Note that, the actual time of computation for the construction of the representations are presented. Codebooks were computed on 1 CPU while PI was constructed by using 32 CPUs in parallel. Moreover, SVM training and prediction time was not taken into consideration. This would further increase computational times especially for PI due to their larger dimension.}
\label{fig:time_score}
\end{figure*}

In Fig.~\ref{fig:time_score}, we plot accuracy vs. time for the proposed approaches and PI for four datasets from EXP-B. We decided to focus on EXP-B here, because it operates on the larger datasets. We vary the codebook size as well as the resolution of PI according to the values provided in Table~\ref{tab:kernelParameters} of the supplementary material. This directly influences the output dimension of the representation and is reflected by the area of the circles in Fig.~\ref{fig:time_score}, i.e. larger diameter means higher dimension. Note that experiments on PBoW were performed on 1 CPU while experiments on PI were performed in parallel on 8 CPUs. Thus the runtime differences are in fact even larger than depicted. For more compact visualization (and avoiding a logarithmic scale which would compress too much) we decided not to take the number of CPUs into account for plotting. We can see clearly that the runtime of PI is always significally larger than for PBoW and sPBoW. The accuracy obtained varies. For all datasets except reddit-12k the performance level of PI is reached (or even superseeded) much quicker. In the case of GeoMat dataset PBoW and sPBoW clearly outperform PI (while consuming much less time) and in case of PetroSurf3D and reddit-5k they quickly achieve a similar performance level. The computational cost of achieving a higher performance with PI e.g. in the case of reddit-12k is over-proportionally high while the performance gain is actually rather limited (approx. $+1\%$). 

% ==============================
\subsection{Qualitative Analysis}

\begin{figure*}
\centering
\includegraphics[width=0.48\linewidth]{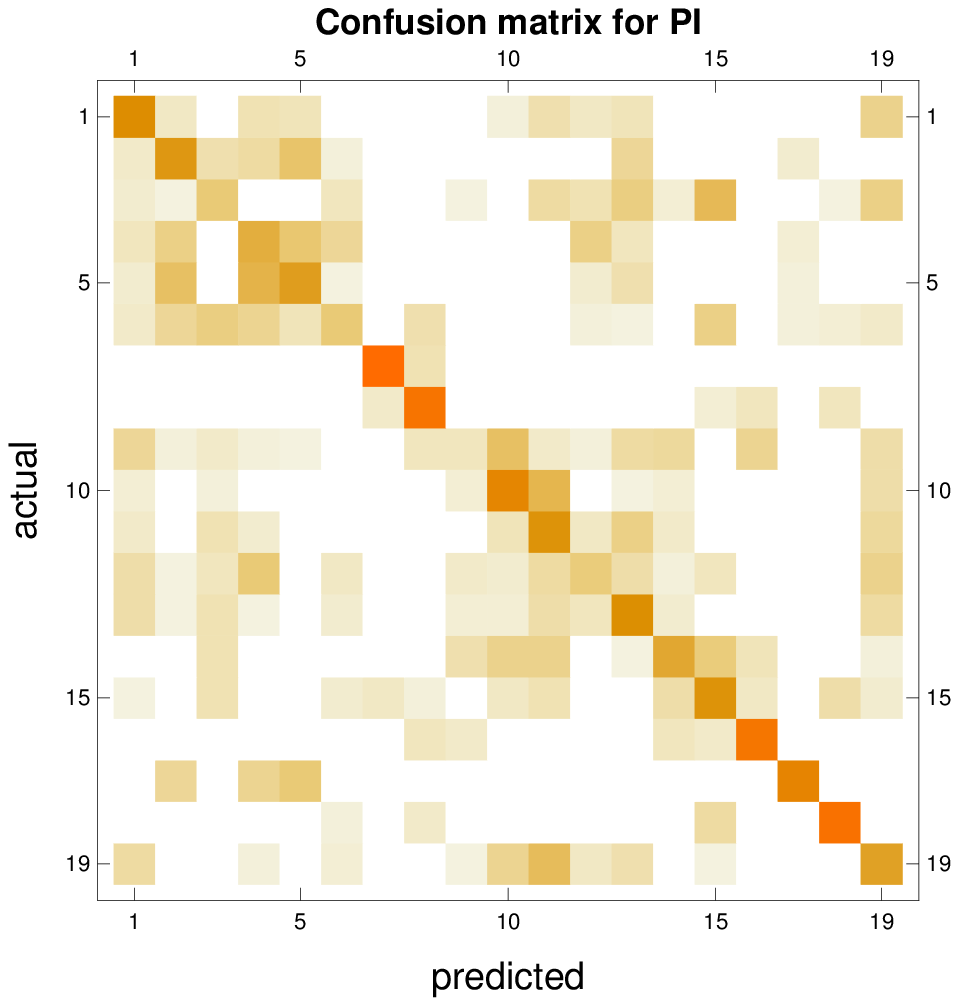}
\includegraphics[width=0.48\linewidth]{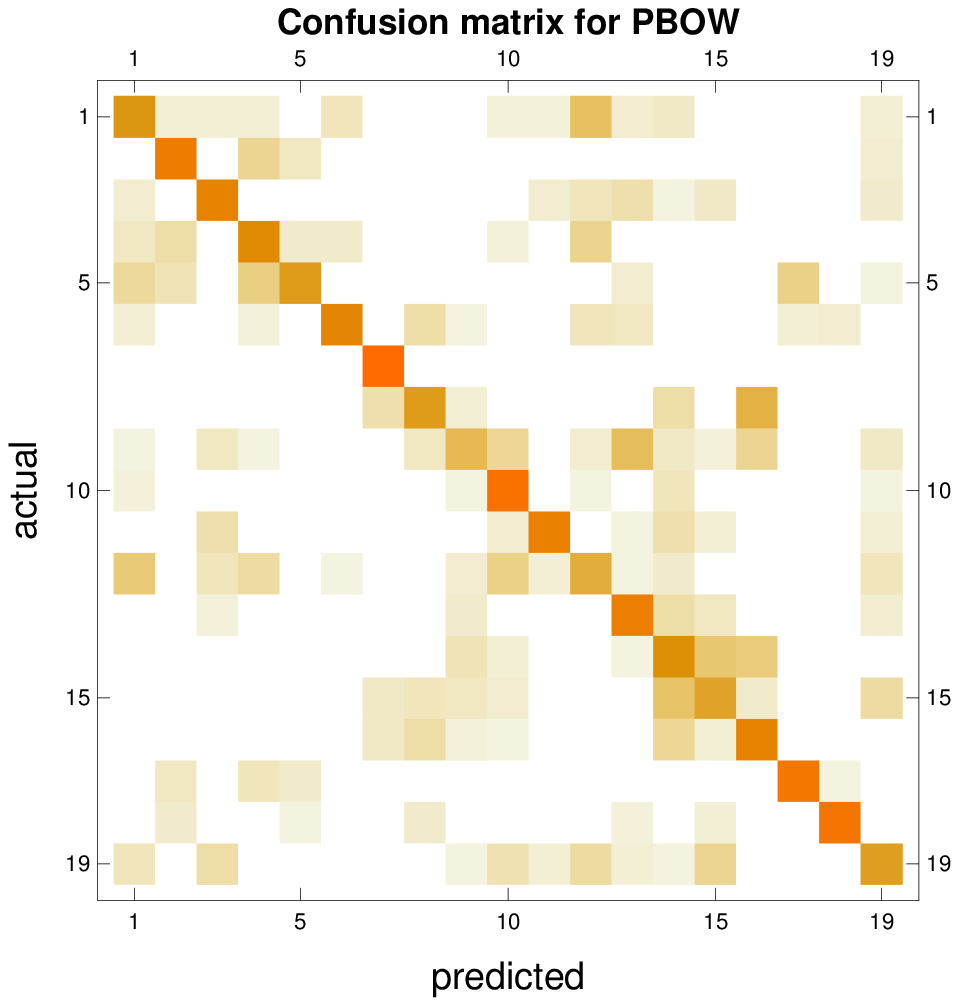}
\caption{Confusion matrix for PI (left) and PBoW (right) on the GeoMat dataset from EXP-A. From the diagonal of the matrices we can see that PBoW outperforms PI for many classes (e.g. classes 2-5, 9 and 12. Furthermore, there are less confusions (off-diagonal values) for PBoW.}
\label{fig:conf_mat_exp3a}
\end{figure*}

\begin{figure*}
\begin{center}
\includegraphics[width=1.\linewidth]{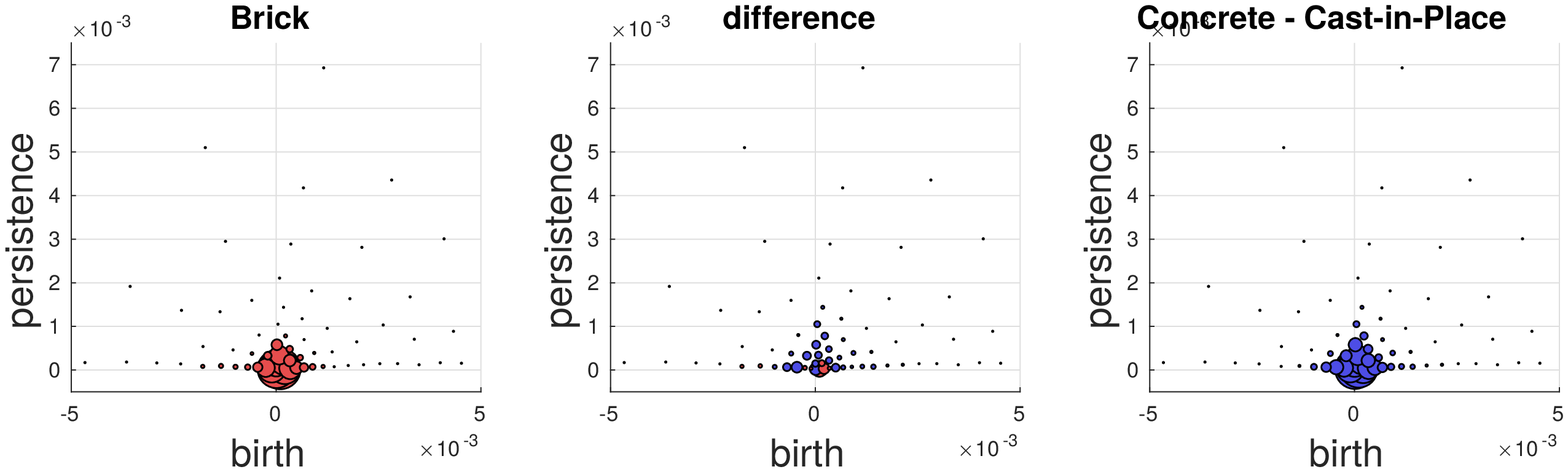}
\includegraphics[width=1.\linewidth]{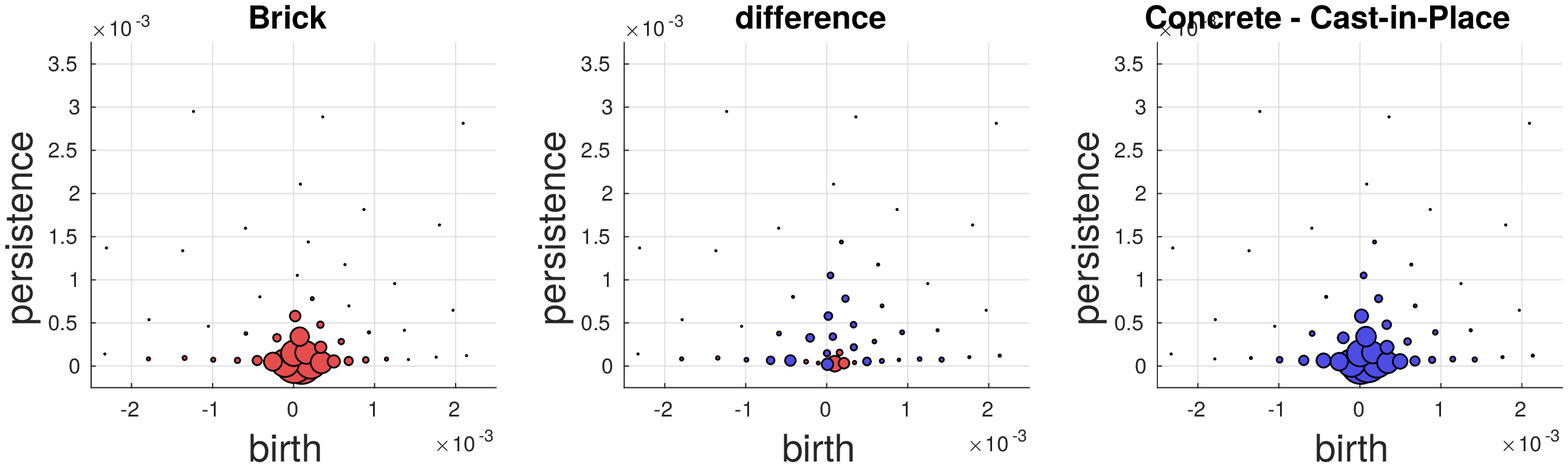}
\includegraphics[width=1.\linewidth]{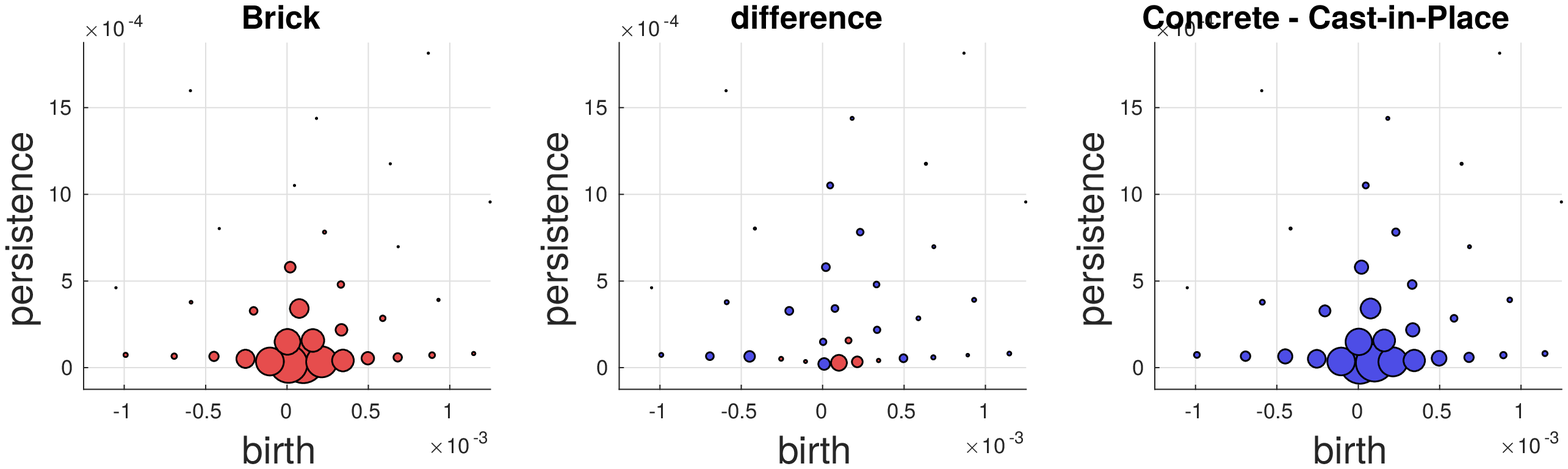}
\caption{Comparison of averaged PBoW histograms for classes ``brick'' and ``concrete cast-in-place'' of the GeoMat dataset (top row: total view, 2nd row: zoomed in view, 3rd row: even further soomed in view). Left and right PDs show the respective histograms in the birth-persistence plane, while in the center we present difference plots between them. Red color in the difference plot means that the left class has stronger support for this cluster and blue means that the right class has stronger support.}
\label{fig:bow_assignment_diff_exp3a_2vs5}
\end{center}
\end{figure*}

In Fig.~\ref{fig:conf_mat_exp3a}, we show the confusion matrix for PI and PBoW on the GeoMat dataset. The matrices show that PBoW achieves a higher number of correct classifications (see the higher values along the diagonal) and less class confusions (lower values for off-diagonal values). A detailed investigation on the discriminative abilities of PBoW between the classes ``cement smooth'' and ``concrete cast-in-place'' (i.e. classes $4$ and $5$) is presented in the paper. In Fig.~\ref{fig:bow_assignment_diff_exp3a_2vs5} we show a similar comparison for classes ``brick'' and ``concrete cast-in-place'' (i.e. classes $2$ and $5$). Also in this example we can observe that spatially fine-grained differences distinguish the classes from each other. The adaptive nature of the bag-of-words model adapts well to such structures.

% ======================================================================
% ======================================================================
{\small
\bibliographystyle{named}
\bibliography{ijcai19}
}